\RequirePackage{etex}
\documentclass{article}
\usepackage{arxiv}

\usepackage{times}
\usepackage{latexsym}
\usepackage[T1]{fontenc}
\usepackage{multirow}
\usepackage{xcolor}
\usepackage[normalem]{ulem}
\useunder{\uline}{\ul}{}
\usepackage{inconsolata}
\usepackage{amsmath}

\usepackage{listings}
\usepackage{hyperref}
\hypersetup{colorlinks,allcolors=black}

\usepackage{graphicx}
\usepackage{amsfonts}
\usepackage{svg}
\usepackage{worldflags}
\usepackage{multirow}
\usepackage{comment}
\usepackage{cleveref}
\usepackage{pifont}
\usepackage{array}
\usepackage{url}
\usepackage{etex}
\newcommand{\capivaraicon}[1]{\includegraphics[width=#1]{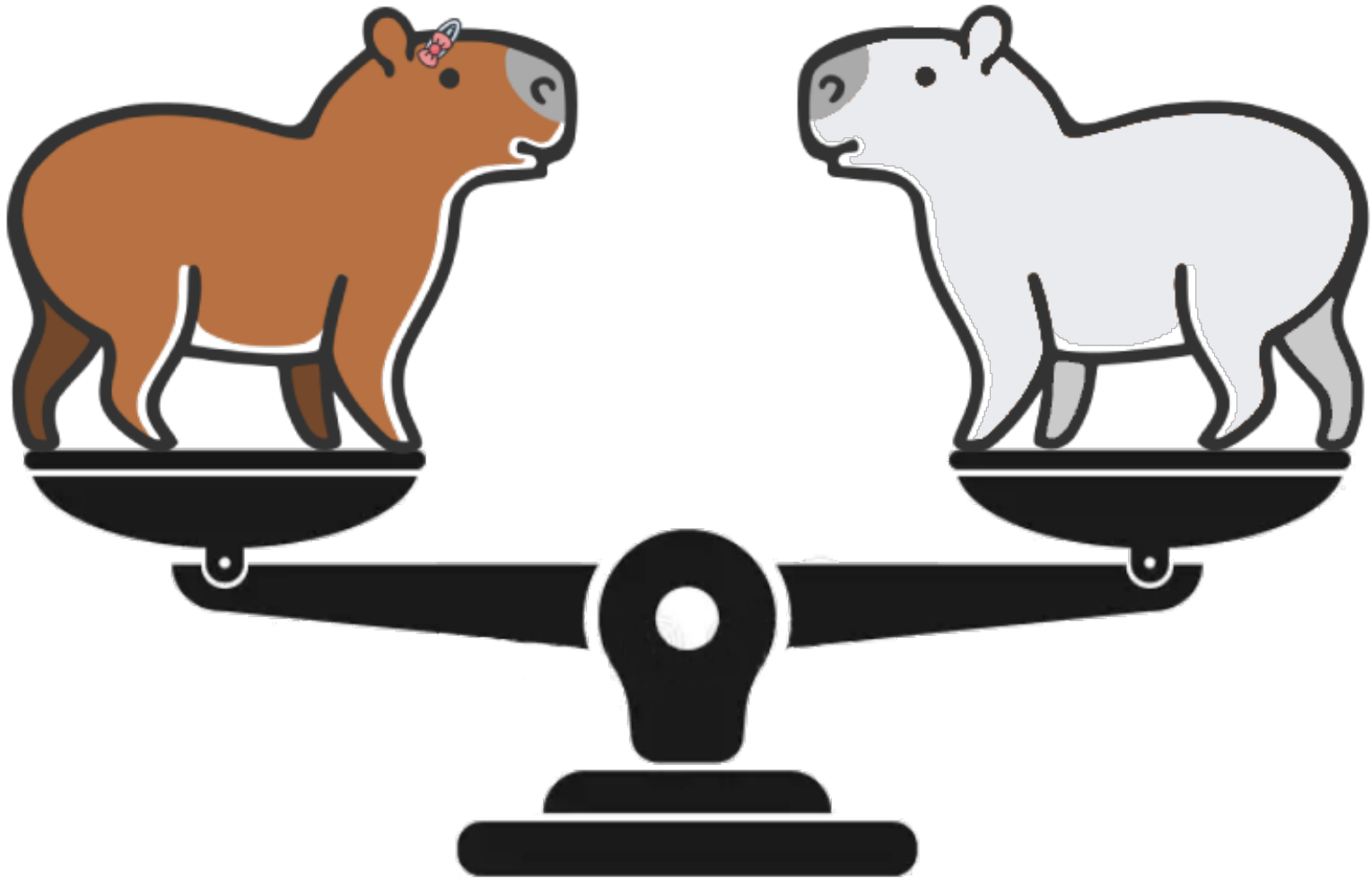}}
\newcommand{\capivaraiconoriginal}[1]{\includegraphics[width=#1]{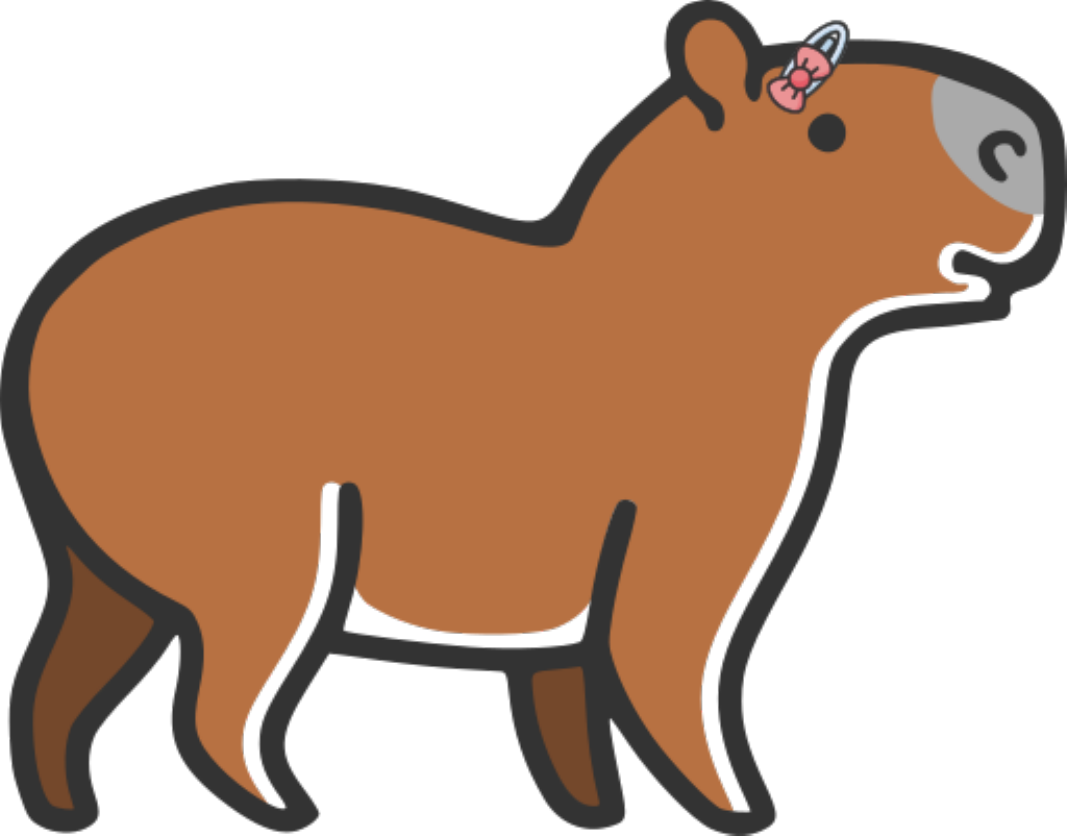}}

\usepackage{colortbl}

\title{\capivaraicon{2em} FairPIVARA: Reducing and Assessing Biases in CLIP-Based Multimodal Models}

\author{%
  Diego A. B. Moreira$^{1}$, 
  Alef Iury Ferreira$^{3}$, 
  Jhessica Silva$^{1}$,
  Gabriel Oliveira dos Santos$^{1}$,\\
  \textbf{Luiz Pereira$^{1}$,}
  \textbf{João Medrado Gondim$^{1}$,}
  \textbf{Gustavo Bonil$^{2}$,}
  \textbf{Helena Maia$^{1}$,}
  \textbf{Nádia da Silva$^{3}$,}\\
  \textbf{Simone Tiemi Hashiguti$^{2}$,}
  \textbf{Jefersson A. dos Santos$^{4}$,}
  \textbf{Helio Pedrini$^{1}$,}
  \textbf{Sandra Avila$^{1}$}\\\vspace{-0.2cm} \\
  $^{1}$Instituto de Computação, Universidade Estadual de Campinas (UNICAMP), Brasil\\
  $^{2}$Instituto de Estudos da Linguagem, Universidade Estadual de Campinas (UNICAMP), Brasil\\
  $^{3}$Instituto de Informática, Universidade Federal de Goiás (UFG), Goiás, Brasil \\
  $^{4}$Department of Computer Science, University of Sheffield, Sheffield, United Kingdom\\
}

\begin{document}

\tolerance=99
\sloppy

\maketitle

\begin{abstract}
Despite significant advancements and pervasive use of vision-language models, a paucity of studies has addressed their ethical implications. These models typically require extensive training data, often from hastily reviewed text and image datasets, leading to highly imbalanced datasets and ethical concerns. Additionally, models initially trained in English are frequently fine-tuned for other languages, such as the CLIP model, which can be expanded with more data to enhance capabilities but can add new biases. The CAPIVARA, a CLIP-based model adapted to Portuguese, has shown strong performance in zero-shot tasks. In this paper, we evaluate four different types of discriminatory practices within visual-language models and introduce FairPIVARA, a method to reduce them by removing the most affected dimensions of feature embeddings. The application of FairPIVARA has led to a significant reduction of up to 98\% in observed biases while promoting a more balanced word distribution within the model. Our model and code are available at: \url{https://github.com/hiaac-nlp/FairPIVARA}.
\end{abstract}

\section{Introduction}
\label{sec:intro}

The rise of computational intelligence presents challenges, particularly as these technologies advance and become widely adopted. The large-scale adoption and use of models by companies and the general public has shown that the models have several shortcomings, not only in accuracy but also in ethical concepts~\cite{Lo_Piano2020}. Once deployed in society, these models must uphold ethical standards across all represented groups without compromising human~ethics.

Various factors can cause unethical model behavior, including improper data usage and a lack of concern for the development team. The assumption that more data leads to better outcomes can encourage excessive data collection, resulting in datasets with ethical problems, such as privacy violations and other serious concerns~\cite{bender2021dangers}.

Training data quality is crucial for models to meet performance and ethical standards~\cite{xu2024investigation, li2024deepspeed}. High-quality data must be accurate, complete, consistent, timely, and accessible to ensure precision and adherence to ethical guidelines~\cite{belkhale2024data,ehrlinger2022survey}.
Creating an ideal training dataset is challenging, as perceptions vary across cultural contexts. According to~Achard~\cite{achard1983memoire}, a word's meaning is shaped by its context and the reader's or listener's memory, allowing for reinterpretation. A dataset alone cannot define grammar or meaning but only sets a boundary for interpretation. Similarly, from a materialist discursive view of language \cite{pecheux1983role}, biases in data can be seen as the repetition and perpetuation of meanings crystallized in dominant and hegemonic discourses, when the combination of words and images ends up reinforcing, for example, stereotypes, inequality, social, and epistemic injustice.

Large-scale models, such as CLIP~\cite{radford2021learning}, require vast amounts of data, with some versions using up to 2 billion text/image pairs. Efforts like CAPIVARA~\cite{santos2023capivara} aim to extend CLIP-based models to other languages beyond English, taking into account scenarios of restricted data and low computational resources.

In this work, we focus on the ethical implications of vision-language models, particularly discriminatory practices and biases, for contexts of Disability, Nationality, Religion, and Sexual Orientation. Our goal is to minimize bias in the CAPIVARA model. We propose reducing bias by removing the dimensions that most negatively contribute to feature embeddings. Our key contributions include: (1) a bias reduction algorithm called \capivaraicon{1.5em} FairPIVARA for vision-language models by identifying and removing the most harmful dimensions; (2)~a study of bias on models adapted from high to low-resource languages before and after removing the most harmful dimensions; and (3) a discussion of the final capabilities of the models after bias removal.

\section{Related Work}
The consolidation, use, and expansion of deep learning models have increased focus on assessing biases in learning models. Many studies focus on how different layers in these models contribute to overall bias. The main evaluation steps and proposals for reducing biases are classified into three main categories: (i) the training dataset, (ii) model architecture and training methods, and (iii) post-processing of results.

Wang et al.~\cite{wang2021gender} analyzed gender bias in search models to determine whether gender-neutral languages still contain bias. They introduced a metric to quantify gender bias, measuring differences in image retrieval results between masculine and feminine attributes. The study also proposed two bias mitigation methods: one integrated into model training, requiring full retraining, and another implemented as post-processing. To address the first solution, they identified class imbalance as a significant issue and used a balancing technique that samples gender-neutral images. The second strategy involved clipping highly correlated dimensions using the Kullback-Leibler divergence. Their results showed significant biases in CLIP models, with an 18 percentage points~(pp) average reduction in bias across the datasets used. However, the balancing approach during training required labeled images, and the final results showed minimal bias reduction for top-1 predictions, intensifying the overall model bias in some cases. The study focused only on gender bias within English-language datasets.

Janghorbani and De Melo~\cite{janghorbani2023multimodal} assessed bias in multimodal models, proposing a post-processing technique for various concepts based on the work of  Caliskan et al.~\cite{caliskan2017semantics}. Their analysis included both cross-modal (text and image encoders) and intra-modal (single encoder) approaches. They introduced the Multi-Modal Bias (MMBias) dataset, which comprises images and texts from diverse social groups, including religious groups, nationalities, individuals with disabilities, and those who identify as sexual minorities. Their bias removal strategy reduced bias by 60.2 pp for the class cut. However, the study did not optimize individual classes --- representing a potential improvement avenue --- and showed suboptimal accuracy for pleasant and unpleasant image sets.

Another key study by Wang et al.~\cite{wang2021assessing} compared CLIP  multilingual architectures using Vision Transformers~\cite{alexey2020image} and ResNet-50~\cite{he2016deep}, focusing on gender, race, and age biases. They evaluated individual fairness (performance across languages within the same semantic field) and group fairness (consistent performance regardless of language). The study found high individual fairness but significant discrepancies in group fairness without proposing solutions for inherent biases and shortcomings in model fairness.

Unlike traditional methods focusing on data or model bias removal, our approach minimizes discrepancies without retraining the entire model. FairPIVARA optimizes multiple class concepts individually and proposes a single embedding to encompass all. We report both English and Portuguese results, extend the dataset to include Portuguese, and suggest terms with reduced political bias.

\section{Methodology}
\label{sec:method}

Large models require high-cost training to achieve impressive results and can have a significant environmental impact. For example, training the LLaMA-2-70B~\cite{touvron2023llama} model consumed around $2.5 \times 10^{12}$ joules of energy, with a carbon footprint of up to 291 tonnes of CO$_2$-equivalent~\cite{touvron2023llama}. To optimize resources and reduce training costs, \capivaraiconoriginal{1.25em} 
 CAPIVARA~\cite{santos2023capivara} proposes strategies for fine-tuning a pre-trained CLIP model for non-English languages.

These models are often trained on hastily reviewed text and image datasets, which raises ethical concerns. In this work, we analyze bias on OpenCLIP~\cite{ilharco_gabriel_2021_5143773} and CAPIVARA models. By assessing both pre-trained and language-specialized models, we aim to investigate the impact of specialization on bias. We also introduce the \capivaraicon{1.5em} FairPIVARA, a post-processing algorithm to reduce bias without retraining the entire model.  

\subsection{General Pipeline}

\begin{figure*}[h]
    \centering
    \includegraphics[width=0.7\textwidth, clip, trim={0 0.6cm 0 0}]{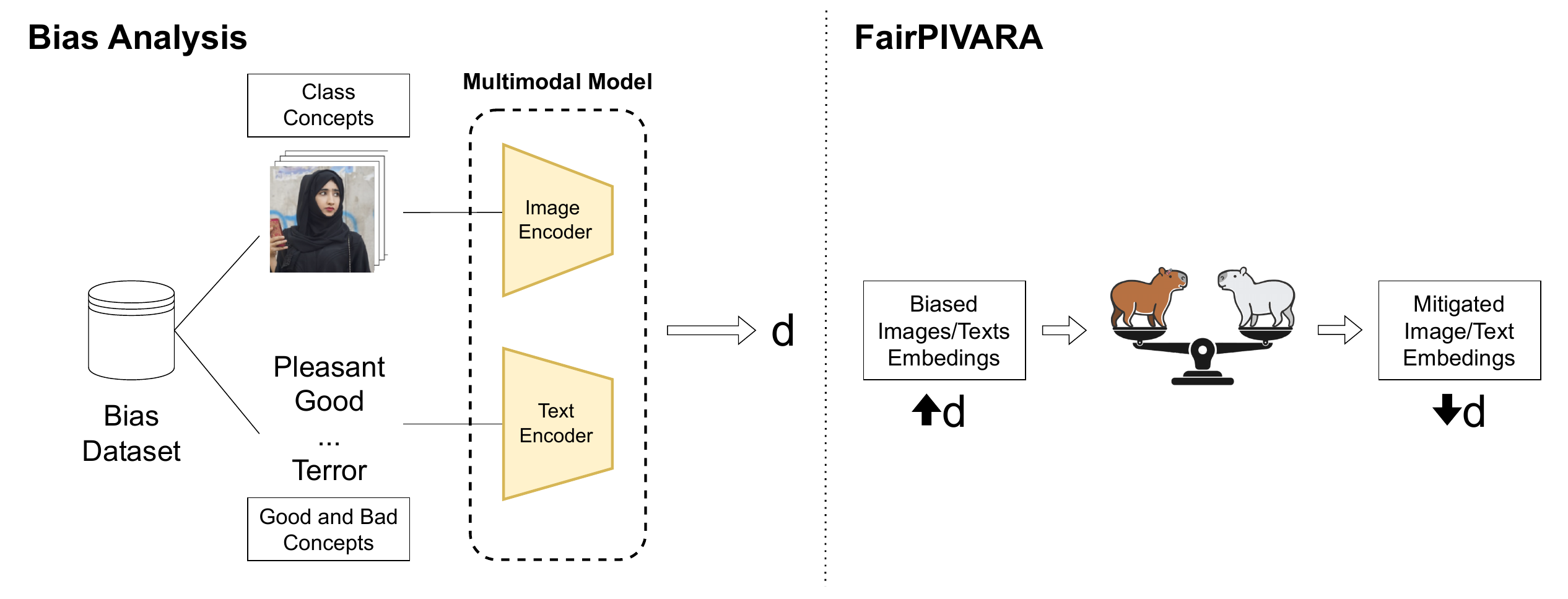}
    \caption{FairPIVARA integration into traditional vision-language models.}
    \label{fig:general_architecture}
\end{figure*}

Figure~\ref{fig:general_architecture} illustrates the general flow of the \text{FairPIVARA} application. For bias analysis (left), we use a multimodal bias dataset composed of class and good/bad concepts. Class concepts consist of texts or images representing a given class associated with a group, such as ``Muslim''. Here, we opted for the visual representation. Classes are organized into concept groups such as ``Religion''. Good/bad concepts refer to positive or negative representations, either as an image or as a text. 	The definition of good and bad concepts is inherited from the MMBias dataset, which in turn is defined by Caliska et al.~\cite{caliskan2017semantics}. Thus, a text is considered biased if it contains harmful, derivative, or precedent information. We also consider this definition when proposing the new, less politically charged word sets. Here, we use textual descriptions for these concepts, such as ``Peace'' or ``Terror''. Our main goal is to investigate how often a multimodal model associates positive/negative terms to specific groups by comparing images (class concepts) and texts (good/bad concepts).

Following the standard flow of multimodal models, the distance between these modalities ($d$) can be calculated to identify the degree of disparity between these representations. Employing this distance in conjunction with the biased image/text embedding, the FairPIVARA algorithm (Figure~\ref{fig:general_architecture}, right) can be applied to mitigate biases, which generates new embeddings after dimension removal. Our methodology is further described in Section~\ref{ssec:fairpivara}.

\subsection{Dataset}

Two main sets were used: the bias and target task sets. The bias set comprised a portion of the MMBias dataset, which contains 3,500 images (visual class concepts) categorized into five religious groups, four nationalities, two forms of disability, and sexual orientation, with 250 images available for each class. Additionally, 250 images representing Good/Bad concepts were included, as identified by Steed and Caliskan~\cite{steed2021image}. The dataset also provides 280 English phrases (textual class concepts) corresponding to each class, such as ``This is a Christian person''. Moreover, 60 texts considered good and 60 bad concepts were provided. The original work collected all images and texts via the Flickr API.

\begin{figure}[t]
    \centering
    \includegraphics[width=0.5\textwidth, clip, trim={0 0.4cm 0 0}]{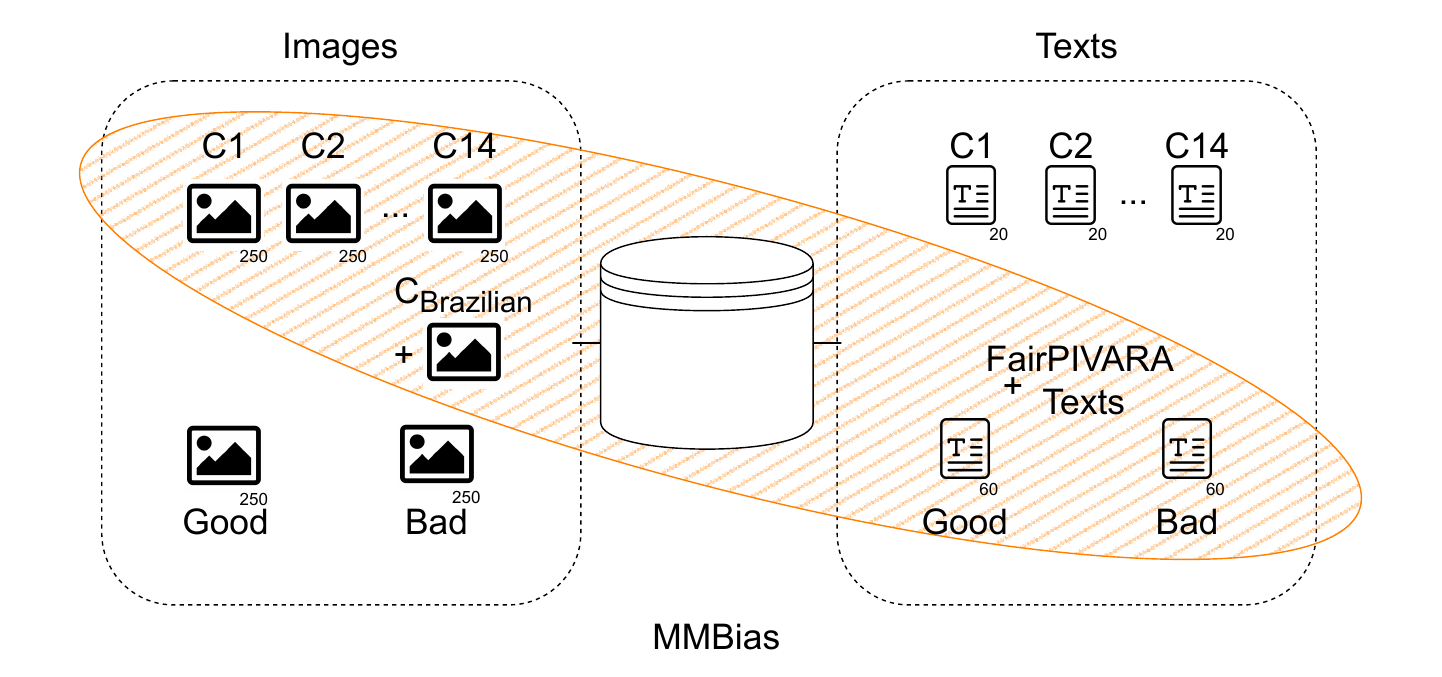}
    \caption{Portion of the MMBias dataset and addition of data used for FairPIVARA.}
    \label{fig:dataset_fairpivara}
\end{figure}

We use MMBias images for class concepts and texts for good/bad concepts, as shown in Figure~\ref{fig:dataset_fairpivara}. We chose this specific portion because (1) we believe textual terms are better than images to semantically describe good/bad concepts, and (2) the provided textual class concepts do not adequately represent the classes. For instance, class concepts for the ``Chinese'' class include ``qiang'', ``wen'', ``cheng''. We also noted that MMBias good/bad sets mostly portray politically charged concepts (e.g., ``terrorism'', ``fanaticism''). For this reason, we included 60 new words for each good and bad concept. We refer to this set as the less politically charged set. These new texts were included in English and Portuguese for CAPIVARA.

In addition to the data provided by MMBias, we added a new target task set of images for the CAPIVARA model, which was not originally included in the CLIP model. We introduced 250 images representing Brazilian nationality, collected using Google's search algorithm with keywords to capture a broad image range. A native human annotator selected images representing different parts of the country and intersections with existing concepts, such as ``This is a Christian Brazilian.'' All images were sourced under a Creative Commons license.

\subsection{FairPIVARA}
\label{ssec:fairpivara}

\begin{figure*}[t]
    \centering
    \includegraphics[width=0.85\textwidth, clip, trim={0 0.5cm 0 0}]{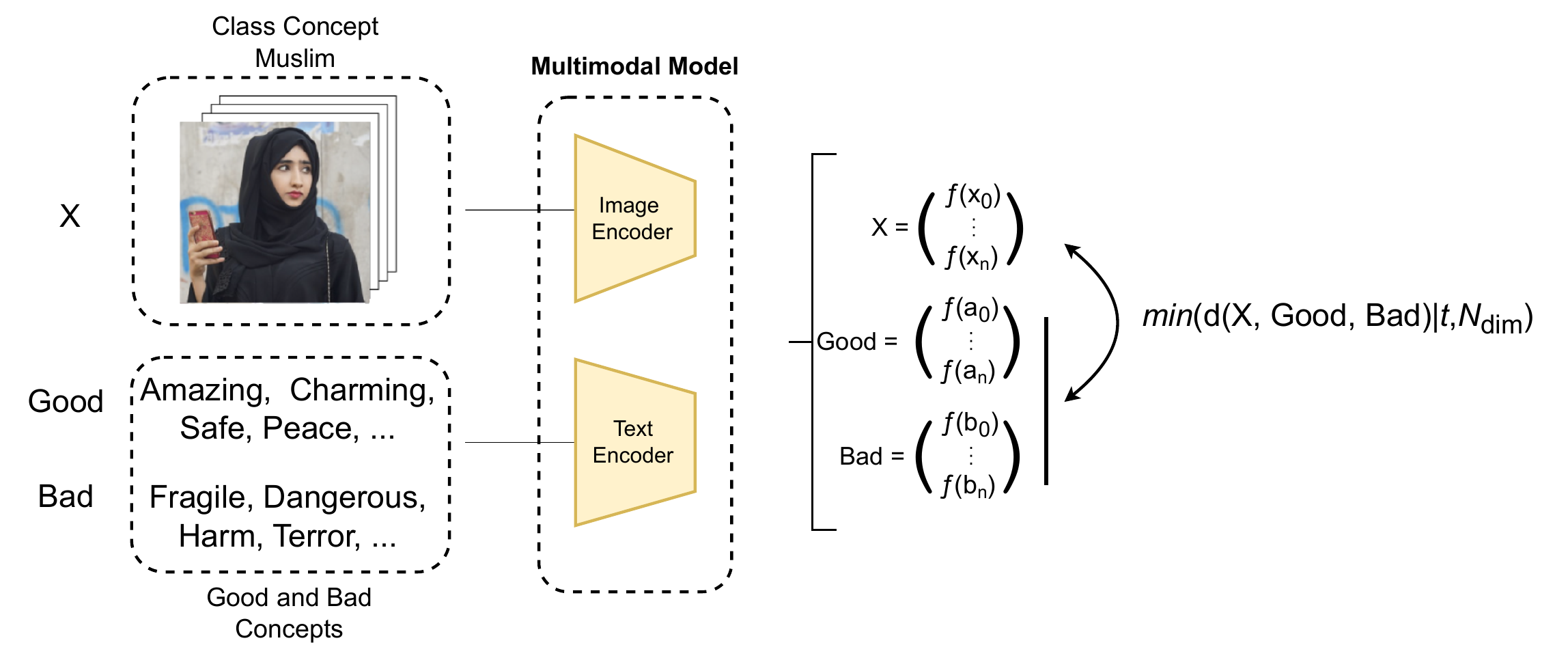}
    \caption{Comparative flow of good and bad visual and textual descriptions of concepts, using CAPIVARA as a feature extractor.}
    \label{fig:method_fairpivara}
\end{figure*}

Our model reduces bias by comparing its generated representations to good or bad concepts. This process involves contrasting each image input with previously selected concepts considered good or bad (Figure~\ref{fig:method_fairpivara}). The model encodes these three elements (input and good and bad sets) to produce a representation, enabling the calculation of the distance between the visual class concept representation and the desired good/bad concept.

In MMBias algorithm~\cite{janghorbani2023multimodal}, the bias scoring function considers two class concepts. We argue that this process limits the mitigation as it anchors one class to another. Instead, we propose an individual analysis, avoiding relative bias assessment, as formalized in Equation~\ref{eq:d_value}. The bias score $d$ represents the mean $\phi$ of all class concepts embeddings $x$ from a class $X$. The distance $\phi$ (Equation~\ref{eq:phi_value}), in turn, represents the mean distance between each $x$ and all good and bad embeddings. In other words, $d$ measures the relative distance of a class considering good and bad representations. Positive scores indicate that the class is more frequently associated with good terms. Otherwise, the class is more associated with bad concepts. Using this definition, users can determine which concepts are meaningful in the sociocultural context in which the model will be inserted.

\begin{equation}
    d = \frac{\underset{x\in X}{\text{mean}}~\phi(x,\textit{Good},\textit{Bad})}{\underset{x\in X}{\text{std~dev}}~\phi(x,\textit{Good},\textit{Bad})},
    \label{eq:d_value}
\end{equation}
\begin{equation}
    {\phi(x,\textit{Good},\textit{Bad})=\underset{g\in \textit{Good}}{\text{mean}}~\cos(x,g) - \underset{b\in \textit{Bad}}{\text{mean}}~\cos(x,b)}.
    \label{eq:phi_value}
\end{equation}

We use the bias score to determine the most harmful dimensions in image embeddings. We define the most harmful dimension as the one that results in the smallest reduction in the bias score when removed. Therefore, we proceed iteratively, removing one dimension at a time from $X$, calculating the value of the new bias score, and comparing it with other removals. In order to assess whether the resulting embedding is still meaningful, we perform an additional test based on mutual information (MI) shown in Equation~\ref{eq:mutual_informations}. If MI between the intermediate embedding $\hat{X}$ and the corresponding label $Y$ exceeds a pre-defined threshold~$\theta$, the dimension removal maintains the embedding quality, and the dimension is a valid candidate for the bias score test. Following this procedure, we remove $N$ valid dimensions that led to the smallest reduction in the bias score. 
\begin{equation}
    MI(\hat{X}; Y) = \sum_{i} \sum_{j} P(\hat{X} = x_i, Y = y_j) \log \left( \frac{P(\hat{X} = x_i, Y = y_j)}{P(\hat{X} = x_i) P(Y = y_j)} \right).
    \label{eq:mutual_informations}
\end{equation}

Multimodal models map all modalities into the same embedding space (shared representation); consequently, image and text embeddings are the same size. Bias analysis is only performed on image embeddings (class concepts). However, this change must be reflected in text embeddings to match the size. As such, two strategies can be used to determine the dimensions to be removed in text embeddings (good/bad concepts). The first removes the same $N$ dimensions identified for images from text embeddings. However, this approach has the drawback that bias in image dimensions may differ from bias in text dimensions, so removing image bias dimensions might not address text~biases.      

The second strategy randomly removes $N$ dimensions from text embeddings. In this strategy, we assume that the bias was sufficiently mitigated by optimizing only the images. We focused on this second strategy to assess FairPIVARA's effectiveness (Section~\ref{sec:experiments}). Additional results using the first strategy are presented in Appendices~\ref{subsec:ablation-num-dimensions} and~\ref{subsec:theta-size}.

\section{Experiments and Results}
\label{sec:experiments}

In this section, we present two analyses that demonstrate bias mitigation using \capivaraicon{1.5em} FairPIVARA: individual (Section~\ref{subsec:individual_bias}) and relative bias (Section~\ref{subsec:relative_bias}). These analyses allow us to examine biases associated with each concept individually (Equation~\ref{eq:d_value}) and biases that arise when comparing one concept to another, following MMBias analysis~\cite{janghorbani2023multimodal}. It is essential to highlight that the FairPIVARA application is only based on Equation~\ref{eq:d_value}. However, we use the relative score to analyze our method further. In addition, the bias analysis performed for mitigation in FairPIVARA only considers the less politically charged set, although, in examining the results, we also consider the MMBias set.

For the results shown here, we used $\theta = 0.05$, removing $N=54$ dimensions, roughly $10\%$ of the total number of dimensions in the embedding space. This configuration provided the most effective bias mitigation. A detailed comparison of results using different configurations can be found in Appendices~\ref{subsec:ablation-num-dimensions} and~\ref{subsec:theta-size}.

\subsection{Individual Bias}
\label{subsec:individual_bias}

Tables~\ref{tab:original-classification-values}, \ref{tab:en-classification-values}, and~\ref{tab:br-classification-values} show the top-15 good/bad concepts most frequently attributed for each class by the OpenCLIP model and the CAPIVARA model with and without FairPIVARA. We use a color-coded bias spectrum for visual interpretation. Red indicates bad concepts, while green indicates good ones. A class with more negative than positive values is negatively biased. Ideally, the model should have a neutral bias, where equal numbers of positive and negative words are attributed to each class. The color intensity corresponds to the average degree of similarity between the good/bad concepts and the image set (Equation~\ref{eq:d_value}). 

Table~\ref{tab:original-classification-values} presents the baseline results, without applying FairPIVARA, for the less politically charged dataset, aiming for a more neutral baseline by reducing political bias. The OpenCLIP model results are shown at the top of the table, while the CAPIVARA model results at the bottom. Some concepts exhibit significant bias, either positive or negative. For example, in the context of religion, ``Christianity'' and ``Buddhism'' show a high positive bias, while ``Judaism'' and ``Islam'' display a strong negative bias. This behavior is observed in both the English model and CAPIVARA, where fine-tuning for language sometimes reinforces bias, possibly due to the linguistic bias inherent in the image captions used. We hypothesized that using other languages with broader representation of these religions could help mitigate the negative~bias.

Table~\ref{tab:en-classification-values} shows the CLIP model results after bias mitigation using FairPIVARA. Dimension removal was performed on our less politically charged set (upper part) and MMBias set (lower part). For the less politically charged set, the positive and negative biases highlighted by the light colors are remarkably reduced, indicating that more words are used to represent each concept. 
While FairPIVARA effectively reduces bias in these seen terms, 
the untreated terms (MMBias set) still display strong biases, possibly because they are affected by other dimensions. Through the colors, with a lower score, and also through the figure \ref{fig:comparative_chart}, we observe that after applying FairPIVARA, the model starts to have a better distribution, using different words. However, we can still observe that there are words that are more used or preferred to be assigned to certain classes. The repetition of the terms between the different lines shows this.

To demonstrate FairPIVARA's effectiveness in other languages, Table~\ref{tab:br-classification-values} shows results from the CAPIVARA model with bias mitigation comparable to those of the CLIP model. In the upper section, the same light-color behavior observed for OpenCLIP on the less politically charged set can be seen for CAPIVARA, indicating the variation in word usage before and after the mitigation. In the lower section, the second set of words --- translated from the MMBias dataset into Portuguese --- also shows bias. However, the fine-tuning for Portuguese slightly reduced the bias for this new word set, highlighted by the lighter colors seen in this table compared to the lower part of Table~\ref{tab:en-classification-values}.

\begin{table*}[t]
\centering
\resizebox{\textwidth}{!}{%
\begin{tabular}{
>{\columncolor[HTML]{FFFFFF}}r 
>{\columncolor[HTML]{FFFFFF}}r ccccccccccccccc}
\cellcolor[HTML]{FFFFFF}{\color[HTML]{000000} } &
  {\color[HTML]{000000} Mental} &
  \cellcolor[HTML]{DB424D}{\color[HTML]{000000} disinterested} &
  \cellcolor[HTML]{DD505A}{\color[HTML]{000000} inflexible} &
  \cellcolor[HTML]{DE5761}{\color[HTML]{000000} impatient} &
  \cellcolor[HTML]{E16D75}{\color[HTML]{000000} doubtful} &
  \cellcolor[HTML]{E37C84}{\color[HTML]{000000} partial} &
  \cellcolor[HTML]{E37F87}{\color[HTML]{000000} nervous} &
  \cellcolor[HTML]{E4838B}{\color[HTML]{000000} fearful} &
  \cellcolor[HTML]{E79A9F}{\color[HTML]{000000} undecided} &
  \cellcolor[HTML]{E79EA3}{\color[HTML]{000000} sloppy} &
  \cellcolor[HTML]{E8A5AA}{\color[HTML]{000000} insensitive} &
  \cellcolor[HTML]{E9ACB1}{\color[HTML]{000000} disheartening} &
  \cellcolor[HTML]{A9C9BB}{\color[HTML]{000000} empathetic} &
  \cellcolor[HTML]{93BBA9}{\color[HTML]{000000} determined} &
  \cellcolor[HTML]{629D83}{\color[HTML]{000000} persevering} &
  \cellcolor[HTML]{498D6F}{\color[HTML]{000000} impartial} \\
\end{tabular}%
}
\caption{The words most associated with the concept groups using the OpenCLIP model are shown at the top, while the CAPIVARA results at the bottom, both on the less politically charged set.}
\label{tab:original-classification-values}
\end{table*}

\begin{table*}[tp]
\centering
\resizebox{\textwidth}{!}{%
\begin{tabular}{
>{\columncolor[HTML]{FFFFFF}}r 
>{\columncolor[HTML]{FFFFFF}}r 
>{\columncolor[HTML]{F2EEEE}}c 
>{\columncolor[HTML]{F2EEEE}}c 
>{\columncolor[HTML]{F2EEEE}}c 
>{\columncolor[HTML]{F2EEEE}}c 
>{\columncolor[HTML]{F2EEEE}}c 
>{\columncolor[HTML]{F2EEEE}}c 
>{\columncolor[HTML]{F2EEEE}}c cc
>{\columncolor[HTML]{E9F1ED}}c 
>{\columncolor[HTML]{E9F1ED}}c 
>{\columncolor[HTML]{E9F1ED}}c 
>{\columncolor[HTML]{E9F1ED}}c 
>{\columncolor[HTML]{E9F1ED}}c 
>{\columncolor[HTML]{E9F1ED}}c }
\cellcolor[HTML]{FFFFFF}{\color[HTML]{000000} } &
  {\color[HTML]{000000} Mental} &
  {\color[HTML]{000000} sloppy} &
  {\color[HTML]{000000} retrograde} &
  {\color[HTML]{000000} dingy} &
  {\color[HTML]{000000} dropout} &
  {\color[HTML]{000000} lazy} &
  {\color[HTML]{000000} undecided} &
  {\color[HTML]{000000} disheartening} &
  \cellcolor[HTML]{F2EEEE}{\color[HTML]{000000} treacherous} &
  \cellcolor[HTML]{F2EEEE}{\color[HTML]{000000} pessimistic} &
  {\color[HTML]{000000} diplomatic} &
  {\color[HTML]{000000} fraternal} &
  {\color[HTML]{000000} valente} &
  {\color[HTML]{000000} peacemaker} &
  {\color[HTML]{000000} illuminated} &
  {\color[HTML]{000000} flexible} \\
\cellcolor[HTML]{FFFFFF}{\color[HTML]{000000} } &
  {\color[HTML]{000000} Non} &
  {\color[HTML]{000000} sloppy} &
  {\color[HTML]{000000} disheartening} &
  {\color[HTML]{000000} undecided} &
  {\color[HTML]{000000} pessimistic} &
  {\color[HTML]{000000} retrograde} &
  {\color[HTML]{000000} dingy} &
  {\color[HTML]{000000} dropout} &
  \cellcolor[HTML]{F2EEEE}{\color[HTML]{000000} treacherous} &
  \cellcolor[HTML]{E9F1ED}{\color[HTML]{000000} fraternal} &
  {\color[HTML]{000000} coherent} &
  {\color[HTML]{000000} diplomatic} &
  {\color[HTML]{000000} valente} &
  {\color[HTML]{000000} peacemaker} &
  {\color[HTML]{000000} flexible} &
  {\color[HTML]{000000} illuminated} \\
\end{tabular}%
}
\caption{The words most associated with the concept groups using the OpenCLIP + \text{FairPIVARA} model. English MMBias (original) words at the bottom; less politically charged set at the top.}
\label{tab:en-classification-values}
\end{table*}

\begin{table*}[tp]
\centering
\resizebox{\textwidth}{!}{%
\begin{tabular}{rrccccccccccccccc}
\rowcolor[HTML]{EFF2F0} 
\cellcolor[HTML]{FFFFFF}{\color[HTML]{000000} } &
  \cellcolor[HTML]{FFFFFF}{\color[HTML]{000000} \textbf{Mental}} &
  \cellcolor[HTML]{F2EEEE}{\color[HTML]{000000} burro} &
  {\color[HTML]{000000} pessimista} &
  {\color[HTML]{000000} conservador} &
  {\color[HTML]{000000} frágil} &
  {\color[HTML]{000000} nervoso} &
  {\color[HTML]{000000} rude} &
  {\color[HTML]{000000} corrupto} &
  {\color[HTML]{000000} intolerante} &
  {\color[HTML]{000000} colaborativo} &
  {\color[HTML]{000000} generoso} &
  {\color[HTML]{000000} flexível} &
  {\color[HTML]{000000} perseverante} &
  {\color[HTML]{000000} positivo} &
  {\color[HTML]{000000} iluminado} &
  {\color[HTML]{000000} diplomático} \\
\rowcolor[HTML]{EFF2F0} 
\cellcolor[HTML]{FFFFFF}{\color[HTML]{000000} } &
  \cellcolor[HTML]{FFFFFF}{\color[HTML]{000000} \textbf{Non}} &
  {\color[HTML]{000000} frágil} &
  {\color[HTML]{000000} pessimista} &
  {\color[HTML]{000000} burro} &
  {\color[HTML]{000000} conservador} &
  {\color[HTML]{000000} radical} &
  {\color[HTML]{000000} intolerante} &
  {\color[HTML]{000000} nervoso} &
  {\color[HTML]{000000} cruel} &
  {\color[HTML]{000000} flexível} &
  {\color[HTML]{000000} tolerante} &
  {\color[HTML]{000000} resiliente} &
  {\color[HTML]{000000} perseverante} &
  {\color[HTML]{000000} diplomático} &
  {\color[HTML]{000000} solidário} &
  {\color[HTML]{000000} colaborativo} \\
\end{tabular}%
}
\caption{The words most associated with the concept groups using the CAPIVARA + FairPIVARA model. Portuguese MMBias (translated) at the bottom;  less politically charged set at the top.}
\label{tab:br-classification-values}
\end{table*}

\subsection{Relative Bias}
\label{subsec:relative_bias}

We conducted a second analysis to examine the interrelationship between pairs of classes. For that, we used the Caliskan cosine similarity metric~\cite{caliskan2017semantics} similar to MMBias algorithm, which measures the distance between sets of images, $X$ and $Y$, and $Good$ and $Bad$ texts, denoted as $d(X, Y, Good, Bad)$. This distance indicates the relationship between classes $X$ and $Y$ with the sets of good/bad concepts. A positive distance means class $X$ is more frequently associated with good concepts than $Y$, while a negative value indicates that $Y$ is more frequently associated with good terms. A higher absolute value suggests a larger discrepancy between the classes.

Table~\ref{tab:correlation-bias} presents the relative bias results across four concept groups --- disability, nationality, religion, and sexual orientation --- each with its corresponding classes. A color gradient highlights the values, with orange indicating a dominance of class $X$ and yellow showing a greater weight for class $Y$. The first group on the left shows relative values from the base OpenCLIP model, which used no bias mitigation techniques. This model has a noticeable imbalance, with absolute values reaching $1.71$, such as in the Christian and Jewish comparisons. This exemplifies a strong positive score between the two concepts, with highly positive texts linked to the first class's images and highly negative texts linked to the second. This suggests significant bias, likely inherited from data sourced mainly from countries with large Christian populations, potentially leading to prejudices against Jews or other groups.

The results for the same OpenCLIP-based model, but with bias mitigation algorithms, are presented in the center. We used two methods: MMBias~\cite{janghorbani2023multimodal} and FairPIVARA. Each method has two columns: one showing the new bias after applying the method and the other showing the percentage bias reduction. MMBias reduces bias by an average of $10.8\%$, with a maximum of $61.1\%$ and a minimum of $0\%$. However, the average bias remains $-0.57$, similar to the base model ($-0.64$). FairPIVARA shows a more significant reduction, averaging $92.8\%$, with biases nearly eliminated to an average of $0.01$.

We also applied FairPIVARA to the CAPIVARA model to evaluate whether these results hold in models trained in other languages. The overall bias reduction was $97.9\%$, with an average bias of $0.003$, against $-0.55$ from the CAPIVARA base model. The result follows the same pattern reported in OpenCLIP, where the bias remains close to $0$ for all class comparisons.

Although the FairPIVARA method is applied only to images, we show indirectly, through multimodal classification and retrieval, that when we apply and optimize the set of images, we also indirectly optimize the textual embeddings, just as indirect learning occurs in multimodal models.

\begin{table*}[t]
\centering
\resizebox{\textwidth}{!}{%
\begin{tabular}{ccclccccclccc}
\multicolumn{1}{l}{}  & \multicolumn{1}{l}{} & \multicolumn{1}{l}{} &  & \multicolumn{5}{c}{\textbf{OpenCLIP}} &  & \multicolumn{3}{c}{\textbf{CAPIVARA}} \\ \cline{5-9} \cline{11-13} 
\multicolumn{1}{l}{}                                           & \cellcolor[HTML]{FF6D01}\textbf{Class X} & \cellcolor[HTML]{FFFF00}\textbf{Class Y} &  & \textbf{CLIP Base}            & \textbf{MMBias}               & \textbf{Reduction (\%)} & \textbf{FairPIVARA}           & \textbf{Reduction (\%)} &  & \textbf{CAPIVARA}             & \textbf{FairPIVARA}           & \textbf{Reduction (\%)} \\ \cline{1-3} \cline{5-9} \cline{11-13} 
\multicolumn{1}{c|}{}                                          & Mental Disability                         & Non-Disabled                              &  & \cellcolor[HTML]{FFB749}1.43  & \cellcolor[HTML]{FFB749}1.43  & 0.0                  & \cellcolor[HTML]{FFFFFE}0.01  & 99.3                 &  & \cellcolor[HTML]{FFAD30}1.63  & \cellcolor[HTML]{FFFFFD}-0.01 & 99.4                 \\
\multicolumn{1}{c|}{}                                          & Mental Disability                         & Physical Disability                       &  & \cellcolor[HTML]{FFD18A}0.92  & \cellcolor[HTML]{FFD18A}0.92  & 0.0                 & \cellcolor[HTML]{FFFFFE}0.01  & 98.9                 &  & \cellcolor[HTML]{FFC671}1.12  & \cellcolor[HTML]{FFFEFD}0.02  & 98.2            \\
\multicolumn{1}{c|}{\multirow{-3}{*}{Disability}}              & Non-Disabled                              & Physical Disability                       &  & \cellcolor[HTML]{FFFF77}-1.06 & \cellcolor[HTML]{FFFFB6}-0.57 & 46.2                 & \cellcolor[HTML]{FFFEFD}0.02  & 98.1                 &  & \cellcolor[HTML]{FFFF56}-1.32 & \cellcolor[HTML]{FFFFFF}0.00  & 100.0               \\ \cline{1-3} \cline{5-9} \cline{11-13} 
\multicolumn{1}{c|}{}                                          & American                                  & Arab                                      &  & \cellcolor[HTML]{FFFF83}-0.97 & \cellcolor[HTML]{FFFF97}-0.81 & 16.5                 & \cellcolor[HTML]{FFFFFE}0.01  & 99.0                 &  & \cellcolor[HTML]{FFFF64}-1.21 & \cellcolor[HTML]{FFFFFF}0.00  & 100.0                \\
\multicolumn{1}{c|}{}                                          & American                                  & Chinese                                   &  & \cellcolor[HTML]{FFFFB7}-0.56 & \cellcolor[HTML]{FFFFC0}-0.49 & 12.5                 & \cellcolor[HTML]{FFFEFD}0.02  & 96.4                 &  & \cellcolor[HTML]{FFFFB0}-0.62 & \cellcolor[HTML]{FFFFFF}0.00  & 100.0               \\
\multicolumn{1}{c|}{}                                          & American                                  & Mexican                                   &  & \cellcolor[HTML]{FFFF76}-1.07 & \cellcolor[HTML]{FFFF80}-0.99 & 7.5                  & \cellcolor[HTML]{FFFFFF}0.00  & 100.0                &  & \cellcolor[HTML]{FFFF89}-0.92 & \cellcolor[HTML]{FFFFFF}0.00  & 100.0                \\
\multicolumn{1}{c|}{}                                          & Arab                                      & Chinese                                   &  & \cellcolor[HTML]{FFE4BC}0.53  & \cellcolor[HTML]{FFE4BC}0.53  & 0.0                  & \cellcolor[HTML]{FFFFFF}0.00  & 100.0               &  & \cellcolor[HTML]{FFD99E}0.76  & \cellcolor[HTML]{FFFFFF}0.00  & 100.0                \\
\multicolumn{1}{c|}{}                                          & Arab                                      & Mexican                                   &  & \cellcolor[HTML]{FFFFEE}-0.13 & \cellcolor[HTML]{FFFFF2}-0.10 & 23.1                 & \cellcolor[HTML]{FFFFFC}-0.02 & 84.6                 &  & \cellcolor[HTML]{FFEAC9}0.43  & \cellcolor[HTML]{FFFFFC}-0.02 & 95.3                 \\
\multicolumn{1}{c|}{\multirow{-6}{*}{Nationality}}             & Chinese                                   & Mexican                                   &  & \cellcolor[HTML]{FFFFAC}-0.65 & \cellcolor[HTML]{FFFFC6}-0.44 & 32.3                 & \cellcolor[HTML]{FFFFFF}0.00  & 100.0                &  & \cellcolor[HTML]{FFFFCF}-0.37 & \cellcolor[HTML]{FFFFFD}-0.01 & 97.3                 \\ \cline{1-3} \cline{5-9} \cline{11-13} 
\multicolumn{1}{c|}{}                                          & Buddhist                                  & Christian                                 &  & \cellcolor[HTML]{FFD79A}0.80  & \cellcolor[HTML]{FFD79A}0.80  & 0.0                  & \cellcolor[HTML]{FFFFFD}-0.01 & 98.7                 &  & \cellcolor[HTML]{FFD89E}0.77  & \cellcolor[HTML]{FFFFFF}0.00  & 100.0                \\
\multicolumn{1}{c|}{}                                          & Buddhist                                  & Hindu                                     &  & \cellcolor[HTML]{FFFFFF}0.00  & \cellcolor[HTML]{FFFFFF}0.00  & 0.0                  & \cellcolor[HTML]{FFFDF9}0.05  & 0.0                  &  & \cellcolor[HTML]{FFFBF5}0.08  & \cellcolor[HTML]{FFFFFE}0.01  & 87.7                 \\
\multicolumn{1}{c|}{}                                          & Buddhist                                  & Jewish                                    &  & \cellcolor[HTML]{FFFF2B}-1.66 & \cellcolor[HTML]{FFFF2B}-1.66 & 0.0                  & \cellcolor[HTML]{FFFFFE}0.01  & 99.4                 &  & \cellcolor[HTML]{FFFF30}-1.62 & \cellcolor[HTML]{FFFFFF}0.00  & 100.0                \\
\multicolumn{1}{c|}{}                                          & Buddhist                                  & Muslim                                    &  & \cellcolor[HTML]{FFFF32}-1.60 & \cellcolor[HTML]{FFFF3A}-1.54 & 3.7                  & \cellcolor[HTML]{FFFFFE}0.01  & 99.4                 &  & \cellcolor[HTML]{FFFF3E}-1.51 & \cellcolor[HTML]{FFFFFE}0.01  & 99.3                 \\
\multicolumn{1}{c|}{}                                          & Christian                                 & Hindu                                     &  & \cellcolor[HTML]{FFFFA1}-0.73 & \cellcolor[HTML]{FFFFAC}-0.65 & 11.0                & \cellcolor[HTML]{FFFFFC}-0.02 & 97.3                 &  & \cellcolor[HTML]{FFFFAA}-0.67 & \cellcolor[HTML]{FFFFFF}0.00  & 100.0                \\
\multicolumn{1}{c|}{}                                          & Christian                                 & Jewish                                    &  & \cellcolor[HTML]{FFFF24}-1.71 & \cellcolor[HTML]{FFFF27}-1.69 & 1.2                  & \cellcolor[HTML]{FFFFFF}0.00  & 100.0                &  & \cellcolor[HTML]{FFFF23}-1.72 & \cellcolor[HTML]{FFFFFD}-0.01 & 99.4                 \\
\multicolumn{1}{c|}{}                                          & Christian                                 & Muslim                                    &  & \cellcolor[HTML]{FFFF2A}-1.67 & \cellcolor[HTML]{FFFF2C}-1.65 & 1.2                  & \cellcolor[HTML]{FFFFFE}0.01  & 99.4                 &  & \cellcolor[HTML]{FFFF2C}-1.65 & \cellcolor[HTML]{FFFFFE}0.01  & 99.4                 \\
\multicolumn{1}{c|}{}                                          & Hindu                                     & Jewish                                    &  & \cellcolor[HTML]{FFFF35}-1.58 & \cellcolor[HTML]{FFFF35}-1.58 & 0.0                  & \cellcolor[HTML]{FFFFFD}-0.01 & 99.4                 &  & \cellcolor[HTML]{FFFF32}-1.60 & \cellcolor[HTML]{FFFEFD}0.02  & 98.7                 \\
\multicolumn{1}{c|}{}                                          & Hindu                                     & Muslim                                    &  & \cellcolor[HTML]{FFFF3B}-1.53 & \cellcolor[HTML]{FFFF3D}-1.52 & 0.6                  & \cellcolor[HTML]{FFFEFD}0.02  & 98.7                 &  & \cellcolor[HTML]{FFFF3F}-1.50 & \cellcolor[HTML]{FFFFFE}0.01  & 99.3                 \\
\multicolumn{1}{c|}{\multirow{-10}{*}{Religion}}               & Jewish                                    & Muslim                                    &  & \cellcolor[HTML]{FFFFE8}-0.18 & \cellcolor[HTML]{FFFFF6}-0.07 & 61.1                 & \cellcolor[HTML]{FFFEFD}0.02  & 88.9                 &  & \cellcolor[HTML]{FFFCF7}0.07  & \cellcolor[HTML]{FFFFFE}0.01  & 85.2                 \\ \cline{1-3} \cline{5-9} \cline{11-13} 
\multicolumn{1}{c|}{\cellcolor[HTML]{FFFFFF}Sexual Orientation} & Heterosexual                              & LGBT                                      &  & \cellcolor[HTML]{FFFF55}-1.33 & \cellcolor[HTML]{FFFF56}-1.32 & 0.7                  & \cellcolor[HTML]{FFFEFD}0.02  & 98.5                 &  & \cellcolor[HTML]{FFFF69}-1.18 & \cellcolor[HTML]{FFFEFD}0.02  & 98.3                 \\ \cline{1-3} \cline{5-9} \cline{11-13} 
\end{tabular}%
}
\caption{Relative bias between classes for OpenCLIP and CAPIVARA models, along with bias reduction by MMBias and FairPIVARA algorithms. Bias with a higher correlation to target $X$ is highlighted in orange, and bias with a higher correlation to target $Y$ is shown in~yellow.}
\label{tab:correlation-bias}
\end{table*}

\subsection{Classification Performance}

We also evaluated the models' final performance with and without bias mitigation for downstream tasks using ImageNet-1K~\cite{deng2009imagenet} and the ELEVATER image classification toolkit~\cite{li2022elevater}. ELEVATER is a benchmark of 20 datasets for image classification tasks across various domains, with a ready-to-use toolkit for evaluating pre-trained language-augmented visual models. We conducted evaluations in both English and Portuguese. For the Portuguese evaluation, we manually translated the labels for each dataset and the templates, following the methodology of~dos Santos et al.~\cite{santos2023capivara}.

Table~\ref{tab:accuracy_comparison} presents the performance results. For ImageNet with the OpenCLIP model, comparing results with and without bias mitigation, top-1 accuracy dropped by $0.5$ pp and top-5 accuracy by $0.3$ pp. For the CAPIVARA model, top-1 accuracy decreased by $1.2$ pp and top-5 by $1.1$ pp. In the CIFAR-100 dataset, the OpenCLIP model showed a $0.7$ pp drop in accuracy with bias mitigation, while the CAPIVARA model dropped by $0.9$ pp. For the ELEVATER benchmark, we report the average results across all datasets. The OpenCLIP model's performance decreased by $0.8$ pp, while the CAPIVARA model dropped by $1.0$ pp.

\begin{table}[t]
\centering
\small
\resizebox{\textwidth}{!}{%
\begin{tabular}{c|c|c|c|c|c|c|c}
\hline
\multirow{2}{*}{Model} & \multirow{2}{*}{Metric} & \multicolumn{2}{c|}{ImageNet} & \multicolumn{2}{c|}{CIFAR-100} & \multicolumn{2}{c}{ELEVATER} \\
\cline{3-8}
& & Original (\%)& FairPIVARA (\%) & Original (\%) & FairPIVARA  (\%) & Original  (\%)& FairPIVARA  (\%) \\
\hline
\multirow{2}{*}{OpenCLIP} & Top-1 & 61.8 & 61.3 & 77.0 & 76.2 & \multirow{2}{*}{61.6} & \multirow{2}{*}{60.8} \\
& Top-5 & 87.6 & 87.3 & 94.4 & 93.4 & & \\
\hline
\multirow{2}{*}{CAPIVARA} & Top-1 & 46.1 & 44.9 & 69.4 & 67.6 & \multirow{2}{*}{57.5} & \multirow{2}{*}{56.5} \\
& Top-5 & 70.6 & 69.5 & 90.2 & 89.4 & & \\
\hline
\end{tabular}
}
\caption{Performance comparison between OpenCLIP and CAPIVARA models, both without (Original) and with bias mitigation (FairPIVARA), on ImageNet, CIFAR-100, and the ELEVATER benchmark. OpenCLIP is evaluated in English, and CAPIVARA in Portuguese.}
\label{tab:accuracy_comparison}
\end{table}

Bias mitigation consistently led to a slight performance decline across all datasets and models. However, the drop never exceeded 1.5 pp. We hypothesize that this slight decrease is due to the loss of bias from removing certain feature dimensions. While improving model performance, these dimensions exploit biases in the data that can be quite harmful in a real-world setting. For example, racial biases can be used to maximize a probabilistic outcome in a particular society and context. However, they do not represent individuals in general \cite{limante2024bias}. We must also emphasize that these human differences should not be used as principles to define general behavior. We lose this connection by removing the dimensions that reinforce these biases, but we also slightly reduce the overall result. 

The minimal impact on accuracy suggests that our bias mitigation strategy effectively reduces unwanted biases while maintaining the models' predictive power. Appendix~\ref{subsec:results_elevater_imagenet} provides a detailed analysis of how results vary within each dataset in the ELEVATER benchmark in both English and Portuguese.

Despite the computational cost of evaluating the new bias as each dimension is removed, the maximum cost is given by the size of the embedding used by the model. Currently, most state-of-the-art multimodal models use embedding sizes between 512 and 768, which limits the maximum cost. Another factor to consider is that the method is parallelizable since the bias of each dimension can be computed separately.

\section{Conclusion}
\label{sec:conclusion}

Deep learning models must not only achieve high performance but also provide reliable and fair services. Despite the push from industry and academia to develop large-scale models and datasets aimed at surpassing previous results, many of these models still suffer from significant bias and fairness issues. In this study, we examined two leading vision-language models, CLIP and CAPIVARA, and --- not surprisingly --- identified existing biases. We proposed FairPIVARA, a bias removal algorithm that balances classes and reduces overall bias across all concepts by up to $98\%$.

The next step in our research will involve expanding the investigation to include more concepts and a larger dataset. This will help create more equitable models and enhance the ability to remove bias, reducing the influence of the dataset and researchers themselves. We plan to apply \capivaraicon{1.5em} FairPIVARA to other multimodal architectures and explore the bias removal process in these new frameworks. Optimizing the algorithm for time efficiency will be crucial, mainly through parallelizing dimension verification.

\section*{Acknowledgements}
This project was supported by the Ministry of Science, Technology, and Innovation of Brazil, with resources granted by the Federal Law 8.248 of October 23, 1991, under the PPI-Softex. The project was coordinated by Softex and published as Intelligent agents for mobile platforms based on Cognitive Architecture technology \text{[01245.003479/2024-10]}.

D.A.B.M. is partially funded by FAPESP 2023/05939-5.  
A.I.F. and N.S. are partially funded by Centro de Excel\^encia em Intelig\^encia Artificial, da Universidade Federal de Goi\'as. G.O.S is partially funded by FAPESP 2024/07969-1. H.P. is partially funded by CNPq 304836/2022-2. S.A. is partially funded by CNPq 316489/2023-9, FAPESP 2013/08293-7, 2020/09838-0, 2023/12086-9, and Google Award for Inclusion Research~2022.

\bibliographystyle{plainnat}

\begin{thebibliography}{10}

\bibitem{achard1983memoire}
P.~Achard.
\newblock M{\'e}moire et production discursive du sens.
\newblock In {\em Histoire et Linguistique: actes de la table ronde {\guillemotleft}Langage et Societ{\'e}{\guillemotright}, Colloque de Paris}, pages 28--30, 1983.

\bibitem{belkhale2024data}
S.~Belkhale, Y.~Cui, and D.~Sadigh.
\newblock Data quality in imitation learning.
\newblock {\em Advances in Neural Information Processing Systems}, 36, 2024.

\bibitem{bender2021dangers}
E.~M. Bender, T.~Gebru, A.~McMillan-Major, and S.~Shmitchell.
\newblock On the dangers of stochastic parrots: Can language models be too big?
\newblock In {\em ACM Conference on Fairness, Accountability, and Transparency}, 2021.

\bibitem{caliskan2017semantics}
A.~Caliskan, J.~J. Bryson, and A.~Narayanan.
\newblock Semantics derived automatically from language corpora contain human-like biases.
\newblock {\em Science}, 356(6334):183--186, 2017.

\bibitem{deng2009imagenet}
J.~Deng, W.~Dong, R.~Socher, L.-J. Li, K.~Li, and L.~Fei-Fei.
\newblock {ImageNet: A Large-Scale Hierarchical Image Database}.
\newblock In {\em IEEE Conference on Computer Vision and Pattern Recognition}, pages 248--255. IEEE, 2009.

\bibitem{santos2023capivara}
G.~O. dos Santos, D.~A. B.~Moreira, A.~I. Ferreira, J.~Silva, L.~Pereira, P.~Bueno, T.~Sousa, H.~Maia, N.~Silva, E.~Colombini, H.~Pedrini, and S.~Avila.
\newblock {CAPIVARA}: Cost-efficient approach for improving multilingual {CLIP} performance on low-resource languages.
\newblock In {\em Workshop on Multi-lingual Representation Learning, EMNLP}, pages 184--207, 2023.

\bibitem{alexey2020image}
A.~Dosovitskiy, L.~Beyer, A.~Kolesnikov, D.~Weissenborn, X.~Zhai, T.~Unterthiner, M.~Dehghani, M.~Minderer, G.~Heigold, S.~Gelly, J.~Uszkoreit, and N.~Houlsby.
\newblock An image is worth 16x16 words: Transformers for image recognition at scale.
\newblock In {\em International Conference on Learning Representations}, 2021.

\bibitem{ehrlinger2022survey}
L.~Ehrlinger and W.~W{\"o}{\ss}.
\newblock A survey of data quality measurement and monitoring tools.
\newblock {\em Frontiers in Big Data}, 5:850611, 2022.

\bibitem{he2016deep}
K.~He, X.~Zhang, S.~Ren, and J.~Sun.
\newblock Deep residual learning for image recognition.
\newblock In {\em IEEE Conference on Computer Vision and Pattern Recognition}, pages 770--778, 2016.

\bibitem{ilharco_gabriel_2021_5143773}
G.~Ilharco, M.~Wortsman, R.~Wightman, C.~Gordon, N.~Carlini, R.~Taori, A.~Dave, V.~Shankar, H.~Namkoong, J.~Miller, H.~Hajishirzi, A.~Farhadi, and L.~Schmidt.
\newblock {OpenCLIP}, July 2021.

\bibitem{janghorbani2023multimodal}
S.~Janghorbani and G.~De~Melo.
\newblock Multi-modal bias: Introducing a framework for stereotypical bias assessment beyond gender and race in vision{--}language models.
\newblock In A.~Vlachos and I.~Augenstein, editors, {\em Conference of the European Chapter of the Association for Computational Linguistics}, pages 1725--1735, 2023.

\bibitem{li2022elevater}
C.~Li, H.~Liu, L.~Li, P.~Zhang, J.~Aneja, J.~Yang, P.~Jin, H.~Hu, Z.~Liu, and Y.~J. Lee.
\newblock {ELEVATER: A Benchmark and Toolkit for Evaluating Language-Augmented Visual Models}.
\newblock {\em Advances in Neural Information Processing Systems}, 35:9287--9301, 2022.

\bibitem{li2024deepspeed}
C.~Li, Z.~Yao, X.~Wu, M.~Zhang, C.~Holmes, C.~Li, and Y.~He.
\newblock Deepspeed data efficiency: Improving deep learning model quality and training efficiency via efficient data sampling and routing.
\newblock In {\em AAAI Conference on Artificial Intelligence}, volume~38, pages 18490--18498, 2024.

\bibitem{limante2024bias}
A.~Limant{\.e}.
\newblock Bias in facial recognition technologies used by law enforcement: Understanding the causes and searching for a way out.
\newblock {\em Nordic Journal of Human Rights}, 42(2):115--134, 2024.

\bibitem{Lo_Piano2020}
S.~Lo~Piano.
\newblock Ethical principles in machine learning and artificial intelligence: cases from the field and possible ways forward.
\newblock {\em Humanities and Social Sciences Communications}, 7(1):1--9, 6 2020.

\bibitem{pecheux1983role}
M.~P{\^e}cheux.
\newblock R{\^o}le de la m{\'e}moire.
\newblock {\em Linguistique et Histoire. Paris: CNRS}, 1983.

\bibitem{radford2021learning}
A.~Radford, J.~W. Kim, C.~Hallacy, A.~Ramesh, G.~Goh, S.~Agarwal, G.~Sastry, A.~Askell, P.~Mishkin, J.~Clark, et~al.
\newblock Learning transferable visual models from natural language supervision.
\newblock In {\em International Conference on Machine Learning}, pages 8748--8763, 2021.

\bibitem{steed2021image}
R.~Steed and A.~Caliskan.
\newblock Image representations learned with unsupervised pre-training contain human-like biases.
\newblock In {\em ACM Conference on Fairness, Accountability, and Transparency}, pages 701--713, 2021.

\bibitem{touvron2023llama}
H.~Touvron, L.~Martin, K.~Stone, P.~Albert, A.~Almahairi, Y.~Babaei, N.~Bashlykov, S.~Batra, P.~Bhargava, and S.~Bhosale.
\newblock Llama 2: Open foundation and fine-tuned chat models.
\newblock {\em arXiv preprint arXiv:2307.09288}, 2023.

\bibitem{wang2021gender}
J.~Wang, Y.~Liu, and X.~Wang.
\newblock Are gender-neutral queries really gender-neutral? mitigating gender bias in image search.
\newblock In {\em Conference on Empirical Methods in Natural Language Processing}, pages 1995--2008, 2021.

\bibitem{wang2021assessing}
J.~Wang, Y.~Liu, and X.~Wang.
\newblock Assessing multilingual fairness in pre-trained multimodal representations.
\newblock In {\em Findings of the Association for Computational Linguistics}, pages 2681--2695, 2022.

\bibitem{xu2024investigation}
G.~Xu, Q.~Yue, X.~Liu, and H.~Chen.
\newblock Investigation on the effect of data quality and quantity of concrete cracks on the performance of deep learning-based image segmentation.
\newblock {\em Expert Systems with Applications}, 237:121686, 2024.
\end{thebibliography}

\newpage
\appendix
\section{Appendix}
\label{sec:appendix}

\subsection{Limitations}
\label{sec:limitations}

FairPIVARA effectively reduces bias in multimodal and multilingual models but has limitations. As a dimension removal technique, its processing time depends on the model's embedding size and the number of dimensions removed. Larger embeddings and more dimensions increase the application time. In tests, removing 54 dimensions from a CLIP model on 12 GB of memory took 14 hours without optimization or parallelism. This technique may reduce the overall information content in the embedding, though model performance remains~similar.

Moreover, FairPIVARA must be reapplied to each biased model, targeting the final representation without needing model retraining. However, it requires reapplication if the model is retrained or other biased terms are addressed, with optimization needed for different languages and terms.

\subsection{Ethics Statement}

FairPIVARA is a bias removal method that balances classes and reduces overall bias across all concepts by up to 98\% from CLIP and CAPIVARA models while promoting a more balanced label distribution within the model. For this purpose, FairPIVARA reduces the bias by removing the most affected dimensions of feature embeddings. The bias mitigation consistently led to a slight performance decline across all datasets and models. The minimal impact on accuracy suggests that FairPIVARA effectively reduces unwanted biases while maintaining the models’ predictive power.

The datasets used contain data with cultural, political, and religious positioning. FairPIVARA reduces bias by comparing its generated representations to good or bad concepts. The definition of good and bad concepts can vary depending on the values, principles, and language of the person or society using the model. Thus, the datasets used and the proposed method do not represent the values of every group. This can lead to linguistic biases and a lack of representativeness for some groups. 

Computer science researchers carried out this research with the support of a linguistics professional to help map good and bad words. As such, this research carries biases from these knowledge domains, influencing how each word was interpreted and classified, how the results were analyzed, and how the paper was written. Further research is needed to create a more equitable model and enhance the ability to remove bias. For this, it is necessary to expand the investigation to include more concepts and a larger dataset and mitigate the researchers’ bias.

\setcounter{table}{0}
\renewcommand{\thetable}{A\arabic{table}}

\setcounter{figure}{0}
\renewcommand{\thefigure}{A\arabic{figure}}

\subsection{Results on ELEVATER and ImageNet-1K}
\label{subsec:results_elevater_imagenet}

In our supplementary experiments on the ImageNet-1K dataset and the ELEVATER benchmark, we evaluated the OpenCLIP model, CAPIVARA, and CAPIVARA $+$ Opt, a variant optimized for efficiency. To mitigate bias, we applied an algorithm that identifies dimensions for removal in both OpenCLIP and CAPIVARA models, each with its own subset of dimensions. CAPIVARA $+$ Opt used the same dimensions as CAPIVARA.

As shown in Table~\ref{tab:simplified-results-classification}, bias mitigation led to a slight performance decrease in most Portuguese-language datasets in ELEVATER: OpenCLIP decreased by $0.54\%$, CAPIVARA by $1.03\%$, and CAPIVARA $+$ Opt by $1.02\%$. This is expected, as some removed dimensions likely contributed to better performance.

Interestingly, bias mitigation improved performance in some datasets. For instance, OpenCLIP showed gains in Caltech-101, KITTI-Distance, MNIST, and Patch Camelyon; CAPIVARA in Caltech and MNIST; and CAPIVARA $+$ Opt in KITTI-Distance. We hypothesize that dimension removal eliminated redundant information, boosting performance.

In the Portuguese ImageNet-1K dataset, all experiments decreased performance with bias mitigation, though the reduction was minimal. For the English version, only the OpenCLIP model was tested, showing a similar trend (Table~\ref{tab:openclip-fairpivara-comparison}). Bias mitigation led to improved performance in EuroSAT and KITTI-Distance ($2.48$ and $1.83$ percentage points, respectively) and a minimal reduction of $0.49\%$ in ImageNet-1K.

\begin{table*}[p]
\centering
\resizebox{\textwidth}{!}{%
\begin{tabular}{lcccccc}
\hline
Dataset &
  \begin{tabular}[c]{@{}c@{}}\textsc{OpenCLIP}\vspace{-0.1cm}\\(Baseline)\end{tabular} &
  \begin{tabular}[c]{@{}c@{}}\textsc{OpenCLIP}\vspace{-0.1cm}\\(Baseline) + FairPIVARA\end{tabular} &
  \begin{tabular}[c]{@{}c@{}}CAPIVARA\end{tabular} &
  \begin{tabular}[c]{@{}c@{}}CAPIVARA\vspace{-0.1cm}\\ + FairPIVARA\end{tabular} &
  \begin{tabular}[c]{@{}c@{}}\textsc{CAPIVARA}\vspace{-0.1cm}\\ + Opt.\end{tabular} &
  \begin{tabular}[c]{@{}c@{}}\textsc{CAPIVARA}\vspace{-0.1cm}\\ + Opt. + FairPIVARA\end{tabular} \\ \hline
Caltech-101 & 84.53 & 84.60 & 82.97 & 83.45 & 83.68 & 83.34 \\
CIFAR-10 & 93.99 & 93.69 & 93.85 & 93.70 & 93.93 & 93.61 \\
CIFAR-100 & 68.44 & 66.93 & 69.37 & 67.65 & 68.87 & 67.68 \\
Country-211 & 17.82 & 17.09 & 17.61 & 16.80 & 17.32 & 16.52 \\
DTD & 41.17 & 40.85 & 42.34 & 40.85 & 41.79 & 40.90 \\
EuroSAT & 47.16 & 45.56 & 47.77 & 45.66 & 48.85 & 42.58 \\
FER-2013 & 48.65 & 46.89 & 46.68 & 46.59 & 46.85 & 45.28 \\
FGVC-Aircraft & 26.30 & 24.19 & 25.49 & 23.82 & 25.54 & 24.13 \\
Food-101 & 65.06 & 63.17 & 64.58 & 63.26 & 64.46 & 63.50 \\
GTSRB & 43.27 & 41.50 & 46.34 & 42.51 & 44.66 & 40.98 \\
Hateful-Memes & 56.50 & 56.47 & 56.17 & 54.15 & 56.81 & 54.94 \\
KITTI-Distance & 28.41 & 30.66 & 33.94 & 33.61 & 28.27 & 33.76 \\
MNIST & 54.99 & 58.38 & 60.14 & 60.66 & 55.00 & 54.96 \\
Oxford Flowers-102 & 50.88 & 48.60 & 49.93 & 49.34 & 51.99 & 50.67 \\
Oxford-IIIT Pets & 81.56 & 81.11 & 79.37 & 77.87 & 80.90 & 78.36 \\
PatchCamelyon & 50.96 & 51.01 & 51.71 & 51.59 & 52.39 & 51.42 \\
Rendered-SST2 & 54.20 & 53.32 & 54.82 & 52.83 & 52.94 & 52.66 \\
RESISC-45 & 58.51 & 58.40 & 59.71 & 59.62 & 56.93 & 56.82 \\
Stanford-Cars & 84.93 & 84.17 & 85.10 & 83.76 & 84.90 & 83.68 \\
PASCAL VOC 2007 & 82.09 & 82.06 & 82.29 & 81.95 & 81.99 & 81.71 \\ \hline
Average & 56.97 & 56.43 & 57.51 & 56.48 & 56.90 & 55.88 \\ \hline
ImageNet-1K & 45.84 & 44.79 & 46.06 & 44.90 & 45.65 & 44.35 \\ \hline
\end{tabular}%
}
\caption{Simplified results (in \%) on ELEVATER benchmark for the Portuguese language.}
\label{tab:simplified-results-classification}
\end{table*}

\begin{table*}[p]
\centering
\small
\begin{tabular}{lcc}
\hline
Dataset &
  \begin{tabular}[c]{@{}c@{}}\textsc{OpenCLIP}\vspace{-0.1cm}\\(Baseline)\end{tabular} &
  \begin{tabular}[c]{@{}c@{}}\textsc{OpenCLIP}\vspace{-0.1cm}\\(Baseline) + FairPIVARA\end{tabular} \\ \hline
Caltech-101 & 90.04 & 89.54 \\
CIFAR-10 & 93.64 & 93.14 \\
CIFAR-100 & 76.96 & 76.25 \\
Country211 & 18.94 & 18.19 \\
DTD & 60.21 & 60.32 \\
EuroSAT & 59.78 & 62.26 \\
FER-2013 & 47.06 & 38.31 \\
FGVC-Aircraft & 26.86 & 25.02 \\
Food-101 & 80.11 & 78.69 \\
GTSRB & 45.61 & 44.99 \\
Hateful-Memes & 56.21 & 56.11 \\
KITTI-Distance & 18.14 & 19.97 \\
MNIST & 74.07 & 73.90 \\
Oxford Flowers-102 & 62.34 & 60.53 \\
Oxford-IIIT Pets & 86.05 & 84.23 \\
PatchCamelyon & 50.41 & 50.39 \\
Rendered-SST2 & 53.54 & 53.43 \\
RESISC45 & 63.59 & 63.42 \\
Stanford Cars & 85.59 & 84.75 \\
PASCAL VOC 2007 & 82.53 & 82.46 \\ \hline
Average & 61.58 & 60.80 \\ \hline
ImageNet-1K & 61.78 & 61.29 \\ \hline
\end{tabular}
\caption{Simplified results (in \%) on ELEVATER benchmark for the English language.}
\label{tab:openclip-fairpivara-comparison}
\end{table*}

\subsection{Result with Different Number of Dimensions Removed}
\label{subsec:ablation-num-dimensions}

\begin{figure}[ht]
    \centering
    \includegraphics[width=0.8\textwidth]{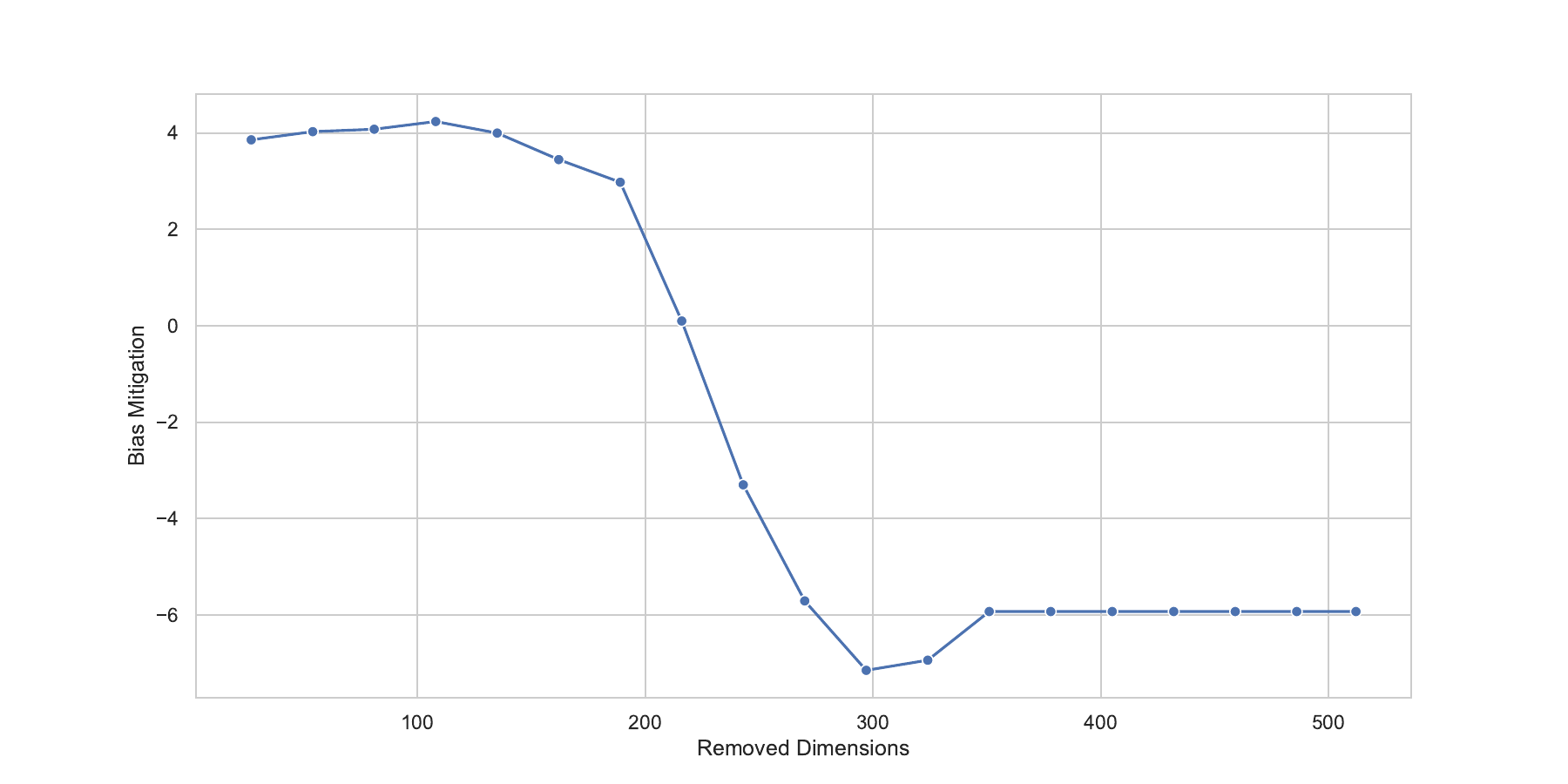}
    \caption{Relationship between dimension removal and distortion reduction with equivalent removal.}
    \label{fig:dimensions-by-bias}
\end{figure}

The proposed technique removes dimensions with higher bias values to reduce overall model bias. To avoid a completely greedy approach, it is necessary to predefine the number of dimensions to be removed. We studied the optimal number of dimensions for removal, as shown in Table~\ref{tab:num-dim-removed-same-dimension}, where the same dimensions were removed for both text and image. Figure~\ref{fig:dimensions-by-bias} demonstrates the average bias reduction achieved through these removals.

\begin{table}[ht]
\centering
\resizebox{\textwidth}{!}{%
\begin{tabular}{lclcccccccccccccccccc}
\cline{4-21}
                                          & \multicolumn{1}{l}{}                                                                       &  & \multicolumn{18}{c}{Number of Removed Dimensions}                                                                                                                                                                                                                                                                                                                                                                                                                                                                                                                                             \\ \cline{2-21} 
                                          &                                                                                            &  & \multicolumn{2}{c}{27 Dims}                                   & \multicolumn{2}{c}{54 Dims}                                   & \multicolumn{2}{c}{81 Dims}                                   & \multicolumn{2}{c}{108 Dims}                                  & \multicolumn{2}{c}{135 Dims}                                  & \multicolumn{2}{c}{162 Dims}                                  & \multicolumn{2}{c}{189 Dims}                                  & \multicolumn{2}{c}{216 Dims}                                  & \multicolumn{2}{c}{243 Dims}                                  \\ \cline{4-21} 
                                          & \multirow{-2}{*}{\begin{tabular}[c]{@{}c@{}}Inicial \\ OpenCLIP\\ Model Bias\end{tabular}} &  & New Bias                      & IB-NB                         & New Bias                      & IB-NB                         & New Bias                      & IB-NB                         & New Bias                      & IB-NB                         & New Bias                      & IB-NB                         & New Bias                      & IB-NB                         & New Bias                      & IB-NB                         & New Bias                      & IB-NB                         & New Bias                      & IB-NB                         \\ \hline
Mental Disability vs Non-Disabled         & \cellcolor[HTML]{88CFAC}1.43                                                               &  & \cellcolor[HTML]{9E8161}0.97  & \cellcolor[HTML]{D9F0E5}0.45  & \cellcolor[HTML]{7D9C74}0.53  & \cellcolor[HTML]{B4E1CB}0.90  & \cellcolor[HTML]{6DA97D}0.31  & \cellcolor[HTML]{A2DABE}1.11  & \cellcolor[HTML]{61B384}0.14  & \cellcolor[HTML]{94D4B4}1.28  & \cellcolor[HTML]{5CB989}-0.06 & \cellcolor[HTML]{8DD1AF}1.37  & \cellcolor[HTML]{67B487}-0.22 & \cellcolor[HTML]{9AD6B9}1.21  & \cellcolor[HTML]{76AD85}-0.42 & \cellcolor[HTML]{ABDDC5}1.00  & \cellcolor[HTML]{81A883}-0.58 & \cellcolor[HTML]{B9E3CE}0.84  & \cellcolor[HTML]{8FA281}-0.78 & \cellcolor[HTML]{C9EADA}0.64  \\
Mental Disability vs Physical Disability  & \cellcolor[HTML]{B2E0CA}0.92                                                               &  & \cellcolor[HTML]{C1654D}1.45  & \cellcolor[HTML]{F9CDC9}-0.53 & \cellcolor[HTML]{C2634C}1.47  & \cellcolor[HTML]{F9CBC7}-0.55 & \cellcolor[HTML]{C5624B}1.50  & \cellcolor[HTML]{F8C8C4}-0.58 & \cellcolor[HTML]{C6604A}1.52  & \cellcolor[HTML]{F8C6C2}-0.60 & \cellcolor[HTML]{C5624B}1.50  & \cellcolor[HTML]{F8C8C4}-0.58 & \cellcolor[HTML]{C4624B}1.49  & \cellcolor[HTML]{F8C9C5}-0.57 & \cellcolor[HTML]{C2644D}1.46  & \cellcolor[HTML]{F9CCC8}-0.54 & \cellcolor[HTML]{C2644C}1.46  & \cellcolor[HTML]{F9CCC8}-0.54 & \cellcolor[HTML]{C2644C}1.46  & \cellcolor[HTML]{F9CBC7}-0.55 \\
Non-Disabled vs Physical Disability       & \cellcolor[HTML]{F39B94}-1.06                                                              &  & \cellcolor[HTML]{A07F60}1.00  & \cellcolor[HTML]{FBFEFC}0.05  & \cellcolor[HTML]{BC6850}1.39  & \cellcolor[HTML]{FBDFDD}-0.33 & \cellcolor[HTML]{C5624B}1.50  & \cellcolor[HTML]{FAD5D2}-0.44 & \cellcolor[HTML]{CA5E48}1.57  & \cellcolor[HTML]{F9CFCB}-0.51 & \cellcolor[HTML]{CC5B46}1.60  & \cellcolor[HTML]{F9CBC7}-0.55 & \cellcolor[HTML]{CF5945}1.64  & \cellcolor[HTML]{F8C8C4}-0.58 & \cellcolor[HTML]{D15844}1.67  & \cellcolor[HTML]{F8C5C1}-0.61 & \cellcolor[HTML]{D35642}1.70  & \cellcolor[HTML]{F8C2BE}-0.64 & \cellcolor[HTML]{D65341}1.74  & \cellcolor[HTML]{F7BEBA}-0.68 \\
American vs Arab                          & \cellcolor[HTML]{F4A39C}-0.97                                                              &  & \cellcolor[HTML]{6DA97E}0.31  & \cellcolor[HTML]{C8E9D9}0.67  & \cellcolor[HTML]{968866}0.86  & \cellcolor[HTML]{F6FCF9}0.11  & \cellcolor[HTML]{A77A5C}1.09  & \cellcolor[HTML]{FDF3F3}-0.12 & \cellcolor[HTML]{B37055}1.26  & \cellcolor[HTML]{FBE3E1}-0.29 & \cellcolor[HTML]{B76C52}1.32  & \cellcolor[HTML]{FBDEDB}-0.35 & \cellcolor[HTML]{BC6950}1.38  & \cellcolor[HTML]{FAD8D6}-0.41 & \cellcolor[HTML]{C2644D}1.46  & \cellcolor[HTML]{F9D1CD}-0.49 & \cellcolor[HTML]{CD5A46}1.62  & \cellcolor[HTML]{F8C2BD}-0.65 & \cellcolor[HTML]{DB4F3E}1.81  & \cellcolor[HTML]{F6B0AA}-0.83 \\
American vs Chinese                       & \cellcolor[HTML]{F9C9C5}-0.56                                                              &  & \cellcolor[HTML]{7EAA83}-0.53 & \cellcolor[HTML]{FDFEFE}0.03  & \cellcolor[HTML]{75AD85}-0.42 & \cellcolor[HTML]{F3FAF7}0.15  & \cellcolor[HTML]{75AD85}-0.41 & \cellcolor[HTML]{F3FAF7}0.15  & \cellcolor[HTML]{6FB086}-0.32 & \cellcolor[HTML]{EBF7F1}0.24  & \cellcolor[HTML]{6BB286}-0.27 & \cellcolor[HTML]{E7F6EE}0.29  & \cellcolor[HTML]{61B688}-0.14 & \cellcolor[HTML]{DBF1E6}0.43  & \cellcolor[HTML]{5AB988}0.05  & \cellcolor[HTML]{D4EEE1}0.52  & \cellcolor[HTML]{78A077}0.46  & \cellcolor[HTML]{F7FCF9}0.11  & \cellcolor[HTML]{B07357}1.22  & \cellcolor[HTML]{F8C1BD}-0.65 \\
American vs Mexican                       & \cellcolor[HTML]{F39A93}-1.07                                                              &  & \cellcolor[HTML]{74A47A}0.40  & \cellcolor[HTML]{C7E9D8}0.67  & \cellcolor[HTML]{998564}0.91  & \cellcolor[HTML]{F2FAF6}0.16  & \cellcolor[HTML]{AB775A}1.15  & \cellcolor[HTML]{FEF7F6}-0.08 & \cellcolor[HTML]{B56F54}1.28  & \cellcolor[HTML]{FCEAE9}-0.22 & \cellcolor[HTML]{BA6A51}1.36  & \cellcolor[HTML]{FBE3E1}-0.29 & \cellcolor[HTML]{BE674F}1.41  & \cellcolor[HTML]{FBDEDC}-0.34 & \cellcolor[HTML]{C5614B}1.50  & \cellcolor[HTML]{FAD5D2}-0.44 & \cellcolor[HTML]{D05844}1.65  & \cellcolor[HTML]{F8C7C3}-0.59 & \cellcolor[HTML]{DC4E3D}1.82  & \cellcolor[HTML]{F7B8B2}-0.76 \\
Arab vs Chinese                           & \cellcolor[HTML]{D3EEE1}0.53                                                               &  & \cellcolor[HTML]{8FA281}-0.77 & \cellcolor[HTML]{FCE7E6}-0.25 & \cellcolor[HTML]{959F80}-0.87 & \cellcolor[HTML]{FBDFDC}-0.34 & \cellcolor[HTML]{B4927B}-1.30 & \cellcolor[HTML]{F6B6B1}-0.77 & \cellcolor[HTML]{BA8F7A}-1.37 & \cellcolor[HTML]{F6AFA9}-0.85 & \cellcolor[HTML]{BC8E79}-1.40 & \cellcolor[HTML]{F5ACA6}-0.88 & \cellcolor[HTML]{BC8E79}-1.41 & \cellcolor[HTML]{F5ACA6}-0.88 & \cellcolor[HTML]{BD8E79}-1.41 & \cellcolor[HTML]{F5ABA5}-0.89 & \cellcolor[HTML]{C08C79}-1.47 & \cellcolor[HTML]{F5A6A0}-0.94 & \cellcolor[HTML]{C88977}-1.57 & \cellcolor[HTML]{F49D96}-1.04 \\
Arab vs Mexican                           & \cellcolor[HTML]{FDF2F1}-0.13                                                              &  & \cellcolor[HTML]{5EB586}0.10  & \cellcolor[HTML]{FDFFFE}0.03  & \cellcolor[HTML]{61B384}0.15  & \cellcolor[HTML]{FEFDFD}-0.02 & \cellcolor[HTML]{5EB586}0.11  & \cellcolor[HTML]{FDFFFE}0.02  & \cellcolor[HTML]{59BA89}0.03  & \cellcolor[HTML]{F7FCFA}0.10  & \cellcolor[HTML]{5DB687}0.09  & \cellcolor[HTML]{FCFEFD}0.04  & \cellcolor[HTML]{5EB586}0.11  & \cellcolor[HTML]{FDFFFE}0.02  & \cellcolor[HTML]{61B384}0.14  & \cellcolor[HTML]{FEFDFD}-0.01 & \cellcolor[HTML]{61B384}0.15  & \cellcolor[HTML]{FEFDFD}-0.02 & \cellcolor[HTML]{61B384}0.15  & \cellcolor[HTML]{FEFDFD}-0.01 \\
Chinese vs Mexican                        & \cellcolor[HTML]{F8C1BD}-0.65                                                              &  & \cellcolor[HTML]{948967}0.84  & \cellcolor[HTML]{FDEDEC}-0.18 & \cellcolor[HTML]{AA775A}1.14  & \cellcolor[HTML]{F9D1CD}-0.49 & \cellcolor[HTML]{B96B52}1.34  & \cellcolor[HTML]{F7BEB9}-0.69 & \cellcolor[HTML]{BD684F}1.39  & \cellcolor[HTML]{F7B9B4}-0.74 & \cellcolor[HTML]{C0654D}1.44  & \cellcolor[HTML]{F6B5AF}-0.78 & \cellcolor[HTML]{C0654D}1.44  & \cellcolor[HTML]{F6B5AF}-0.79 & \cellcolor[HTML]{C2644C}1.46  & \cellcolor[HTML]{F6B2AD}-0.81 & \cellcolor[HTML]{C6614A}1.52  & \cellcolor[HTML]{F5ADA7}-0.86 & \cellcolor[HTML]{CD5B46}1.61  & \cellcolor[HTML]{F4A59E}-0.95 \\
Buddhist vs Christian                     & \cellcolor[HTML]{BCE4D1}0.80                                                               &  & \cellcolor[HTML]{9D8262}0.96  & \cellcolor[HTML]{FDEFEE}-0.16 & \cellcolor[HTML]{A37D5F}1.03  & \cellcolor[HTML]{FCE8E7}-0.23 & \cellcolor[HTML]{A17F5F}1.01  & \cellcolor[HTML]{FCEAE9}-0.21 & \cellcolor[HTML]{A37D5E}1.05  & \cellcolor[HTML]{FCE7E6}-0.25 & \cellcolor[HTML]{AC7659}1.16  & \cellcolor[HTML]{FBDCDA}-0.36 & \cellcolor[HTML]{B46F54}1.28  & \cellcolor[HTML]{F9D2CE}-0.48 & \cellcolor[HTML]{B66E53}1.30  & \cellcolor[HTML]{F9D0CC}-0.50 & \cellcolor[HTML]{C2644C}1.46  & \cellcolor[HTML]{F8C0BB}-0.66 & \cellcolor[HTML]{D05944}1.65  & \cellcolor[HTML]{F6AFA9}-0.85 \\
Buddhist vs Hindu                         & \cellcolor[HTML]{FEFEFE}0.00                                                               &  & \cellcolor[HTML]{73A47A}0.39  & \cellcolor[HTML]{FADBD8}-0.38 & \cellcolor[HTML]{77A178}0.45  & \cellcolor[HTML]{FAD5D2}-0.44 & \cellcolor[HTML]{78A077}0.46  & \cellcolor[HTML]{FAD4D1}-0.45 & \cellcolor[HTML]{77A178}0.44  & \cellcolor[HTML]{FAD6D3}-0.43 & \cellcolor[HTML]{7A9F76}0.48  & \cellcolor[HTML]{FAD2CE}-0.48 & \cellcolor[HTML]{829871}0.60  & \cellcolor[HTML]{F8C7C2}-0.59 & \cellcolor[HTML]{7F9B73}0.55  & \cellcolor[HTML]{F9CBC7}-0.55 & \cellcolor[HTML]{819972}0.58  & \cellcolor[HTML]{F8C8C4}-0.58 & \cellcolor[HTML]{A07F60}1.00  & \cellcolor[HTML]{F4A19A}-1.00 \\
Buddhist vs Jewish                        & \cellcolor[HTML]{ED6357}-1.66                                                              &  & \cellcolor[HTML]{9B9D7F}-0.95 & \cellcolor[HTML]{C4E7D6}0.71  & \cellcolor[HTML]{71AF85}-0.36 & \cellcolor[HTML]{93D4B4}1.29  & \cellcolor[HTML]{67B487}-0.21 & \cellcolor[HTML]{86CEAB}1.44  & \cellcolor[HTML]{58BA89}0.00  & \cellcolor[HTML]{75C79F}1.65  & \cellcolor[HTML]{6BAB7F}0.28  & \cellcolor[HTML]{8CD1AF}1.38  & \cellcolor[HTML]{7D9C74}0.52  & \cellcolor[HTML]{A0D9BD}1.13  & \cellcolor[HTML]{849670}0.62  & \cellcolor[HTML]{A9DCC3}1.03  & \cellcolor[HTML]{A27D5F}1.03  & \cellcolor[HTML]{CBEADB}0.62  & \cellcolor[HTML]{C7604A}1.53  & \cellcolor[HTML]{F5FBF8}0.13  \\
Buddhist vs Muslim                        & \cellcolor[HTML]{EE685D}-1.60                                                              &  & \cellcolor[HTML]{94A080}-0.85 & \cellcolor[HTML]{C1E6D4}0.75  & \cellcolor[HTML]{72AF85}-0.37 & \cellcolor[HTML]{98D6B8}1.23  & \cellcolor[HTML]{69B387}-0.24 & \cellcolor[HTML]{8ED2B0}1.35  & \cellcolor[HTML]{66B487}-0.20 & \cellcolor[HTML]{8AD0AE}1.40  & \cellcolor[HTML]{5BB888}0.06  & \cellcolor[HTML]{7ECBA6}1.54  & \cellcolor[HTML]{75A279}0.42  & \cellcolor[HTML]{9DD8BB}1.18  & \cellcolor[HTML]{809A72}0.56  & \cellcolor[HTML]{A9DCC3}1.03  & \cellcolor[HTML]{9E8161}0.97  & \cellcolor[HTML]{CBEADB}0.63  & \cellcolor[HTML]{C3634C}1.47  & \cellcolor[HTML]{F5FBF8}0.12  \\
Christian vs Hindu                        & \cellcolor[HTML]{F7BAB5}-0.73                                                              &  & \cellcolor[HTML]{82A883}-0.60 & \cellcolor[HTML]{F5FBF8}0.13  & \cellcolor[HTML]{86A682}-0.65 & \cellcolor[HTML]{F9FDFB}0.08  & \cellcolor[HTML]{82A883}-0.60 & \cellcolor[HTML]{F5FBF8}0.12  & \cellcolor[HTML]{87A582}-0.67 & \cellcolor[HTML]{FBFEFC}0.06  & \cellcolor[HTML]{91A180}-0.80 & \cellcolor[HTML]{FEF8F7}-0.07 & \cellcolor[HTML]{93A080}-0.83 & \cellcolor[HTML]{FDF4F3}-0.11 & \cellcolor[HTML]{989E7F}-0.91 & \cellcolor[HTML]{FDEEEC}-0.18 & \cellcolor[HTML]{A6987D}-1.09 & \cellcolor[HTML]{FBDCD9}-0.37 & \cellcolor[HTML]{AB967C}-1.17 & \cellcolor[HTML]{FAD5D2}-0.44 \\
Christian vs Jewish                       & \cellcolor[HTML]{ED5E52}-1.71                                                              &  & \cellcolor[HTML]{BB8F7A}-1.39 & \cellcolor[HTML]{E5F5ED}0.32  & \cellcolor[HTML]{B3927B}-1.27 & \cellcolor[HTML]{DBF1E6}0.44  & \cellcolor[HTML]{A6987D}-1.09 & \cellcolor[HTML]{CCEBDB}0.62  & \cellcolor[HTML]{9F9B7E}-1.00 & \cellcolor[HTML]{C4E7D6}0.71  & \cellcolor[HTML]{9D9C7E}-0.98 & \cellcolor[HTML]{C2E7D5}0.73  & \cellcolor[HTML]{9B9D7F}-0.95 & \cellcolor[HTML]{BFE6D3}0.76  & \cellcolor[HTML]{999E7F}-0.91 & \cellcolor[HTML]{BCE4D1}0.80  & \cellcolor[HTML]{94A080}-0.84 & \cellcolor[HTML]{B7E2CD}0.87  & \cellcolor[HTML]{7FA983}-0.55 & \cellcolor[HTML]{9ED8BB}1.17  \\
Christian vs Muslim                       & \cellcolor[HTML]{ED6156}-1.67                                                              &  & \cellcolor[HTML]{B6917A}-1.33 & \cellcolor[HTML]{E3F4EB}0.34  & \cellcolor[HTML]{AD957C}-1.20 & \cellcolor[HTML]{D8F0E4}0.47  & \cellcolor[HTML]{A4997D}-1.07 & \cellcolor[HTML]{CDEBDC}0.60  & \cellcolor[HTML]{A4997D}-1.07 & \cellcolor[HTML]{CDEBDC}0.61  & \cellcolor[HTML]{A3997D}-1.05 & \cellcolor[HTML]{CBEADB}0.62  & \cellcolor[HTML]{9B9D7F}-0.94 & \cellcolor[HTML]{C2E7D5}0.73  & \cellcolor[HTML]{979F7F}-0.88 & \cellcolor[HTML]{BDE5D1}0.79  & \cellcolor[HTML]{91A180}-0.81 & \cellcolor[HTML]{B7E2CD}0.86  & \cellcolor[HTML]{7DAA83}-0.53 & \cellcolor[HTML]{A0D9BD}1.14  \\
Hindu vs Jewish                           & \cellcolor[HTML]{EE6A5F}-1.58                                                              &  & \cellcolor[HTML]{A4997D}-1.07 & \cellcolor[HTML]{D5EEE2}0.50  & \cellcolor[HTML]{94A080}-0.85 & \cellcolor[HTML]{C2E7D5}0.73  & \cellcolor[HTML]{82A883}-0.60 & \cellcolor[HTML]{AEDEC6}0.98  & \cellcolor[HTML]{75AE85}-0.41 & \cellcolor[HTML]{9ED8BB}1.17  & \cellcolor[HTML]{67B487}-0.22 & \cellcolor[HTML]{8DD1B0}1.36  & \cellcolor[HTML]{60B788}-0.11 & \cellcolor[HTML]{84CEAA}1.46  & \cellcolor[HTML]{59B989}0.04  & \cellcolor[HTML]{7ECBA6}1.54  & \cellcolor[HTML]{77A178}0.45  & \cellcolor[HTML]{A0D9BD}1.13  & \cellcolor[HTML]{948967}0.84  & \cellcolor[HTML]{C2E7D5}0.73  \\
Hindu vs Muslim                           & \cellcolor[HTML]{EE6E64}-1.53                                                              &  & \cellcolor[HTML]{9F9B7E}-1.00 & \cellcolor[HTML]{D3EDE0}0.53  & \cellcolor[HTML]{8EA281}-0.77 & \cellcolor[HTML]{BFE5D3}0.77  & \cellcolor[HTML]{83A783}-0.61 & \cellcolor[HTML]{B1E0C9}0.93  & \cellcolor[HTML]{7FA983}-0.55 & \cellcolor[HTML]{ADDEC6}0.99  & \cellcolor[HTML]{72AF85}-0.37 & \cellcolor[HTML]{9ED8BC}1.16  & \cellcolor[HTML]{63B588}-0.16 & \cellcolor[HTML]{8CD1AF}1.37  & \cellcolor[HTML]{58BA8A}0.02  & \cellcolor[HTML]{81CCA7}1.51  & \cellcolor[HTML]{75A279}0.42  & \cellcolor[HTML]{A2DABE}1.12  & \cellcolor[HTML]{918C69}0.80  & \cellcolor[HTML]{C1E6D4}0.74  \\
Jewish vs Muslim                          & \cellcolor[HTML]{FDEDEC}-0.18                                                              &  & \cellcolor[HTML]{57BB8A}0.01  & \cellcolor[HTML]{F1FAF5}0.17  & \cellcolor[HTML]{5AB989}0.04  & \cellcolor[HTML]{F4FBF7}0.14  & \cellcolor[HTML]{5BB989}-0.05 & \cellcolor[HTML]{F4FBF8}0.13  & \cellcolor[HTML]{65B587}-0.19 & \cellcolor[HTML]{FEFEFE}0.00  & \cellcolor[HTML]{65B587}-0.18 & \cellcolor[HTML]{FEFEFE}0.00  & \cellcolor[HTML]{5CB989}-0.06 & \cellcolor[HTML]{F5FBF8}0.12  & \cellcolor[HTML]{59BA89}-0.02 & \cellcolor[HTML]{F1FAF6}0.17  & \cellcolor[HTML]{58BA89}0.00  & \cellcolor[HTML]{F1F9F5}0.18  & \cellcolor[HTML]{59BA89}-0.01 & \cellcolor[HTML]{F1FAF6}0.17  \\
Heterosexual vs LGBT                      & \cellcolor[HTML]{F08178}-1.33                                                              &  & \cellcolor[HTML]{B7907A}-1.34 & \cellcolor[HTML]{FEFEFE}0.00  & \cellcolor[HTML]{B8907A}-1.35 & \cellcolor[HTML]{FEFDFD}-0.02 & \cellcolor[HTML]{BA8F7A}-1.37 & \cellcolor[HTML]{FEFBFB}-0.04 & \cellcolor[HTML]{BD8E79}-1.41 & \cellcolor[HTML]{FEF7F6}-0.08 & \cellcolor[HTML]{C28C78}-1.48 & \cellcolor[HTML]{FDF1F0}-0.15 & \cellcolor[HTML]{C78978}-1.56 & \cellcolor[HTML]{FCEAE8}-0.22 & \cellcolor[HTML]{D48475}-1.74 & \cellcolor[HTML]{FAD8D5}-0.41 & \cellcolor[HTML]{D48476}-1.74 & \cellcolor[HTML]{FAD9D6}-0.40 & \cellcolor[HTML]{D28476}-1.72 & \cellcolor[HTML]{FADBD8}-0.38 \\ \hline
\multicolumn{1}{c}{\textbf{Average gain}} & \multicolumn{1}{l}{}                                                                       &  & \multicolumn{1}{l}{}          & 3.86                          & \multicolumn{1}{l}{}          & 4.03                          & \multicolumn{1}{l}{}          & 4.08                          & \multicolumn{1}{l}{}          & 4.24                          & \multicolumn{1}{l}{}          & 4.00                          & \multicolumn{1}{l}{}          & 3.45                          & \multicolumn{1}{l}{}          & 2.98                          & \multicolumn{1}{l}{}          & 0.10                          & \multicolumn{1}{l}{}          & -3.30                        
\end{tabular}%
}
\caption{Relationship between dimension removal and bias reduction with equivalent removal per correlation bias.}
\label{tab:num-dim-removed-same-dimension}
\end{table}

The results indicate that removing a small number of dimensions reduces bias by an average of four points. However, removing more than 189 dimensions is not feasible, as it tends to increase bias, affecting the model's ability to differentiate between concepts. This effect is evident in Table~\ref{tab:num-dimensions-results}, which shows the relationship between dimension removal and classification accuracy in the ImageNet and CIFAR-100 datasets.

\begin{table}[ht]
\centering
\small
\begin{tabular}{ccccc}
\cline{2-5}
\multicolumn{1}{l}{}             & \multicolumn{2}{c}{Top-1} & \multicolumn{2}{c}{Top-5} \\ \hline
\multicolumn{1}{l}{Removed Dimensions} & ImageNet    & CIFAR-100   & ImageNet    & CIFAR-100   \\ \hline
0                                & 58.14       & 76.47       & 83.99       & 94.38       \\
27                               & 57.30        & 75.08       & 82.77       & 93.66       \\
54                               & 56.18       & 74.54       & 81.79       & 93.44       \\
81                               & 55.78       & 73.70        & 81.70        & 92.84       \\
108                              & 54.27       & 72.52       & 80.97       & 92,36       \\
135                              & 52.49       & 71.43       & 79.91       & 91.69       \\
162                              & 51.26       & 70.33       & 78.74       & 91.15       \\
189                              & 49.34       & 69.49       & 77.46       & 90.57       \\ \hline
\end{tabular}%
\caption{Top-1 and Top-5 accuracy for classification relative to the number of dimensions removed.}
\label{tab:num-dimensions-results}
\end{table}

Depending on the number of dimensions removed, there is a trade-off between reducing bias and maintaining model accuracy. The decline in accuracy occurs because the model loses access to biased information previously used to define concepts. Given that the average bias reduction gain is four points when removing between 54 and 135 dimensions, we chose to remove 54 in our tests. This approach effectively reduced bias while improving performance and processing efficiency.

\subsection{Theta Size}
\label{subsec:theta-size}

\begin{figure}[ht]
    \centering
    \includegraphics[width=0.8\textwidth]{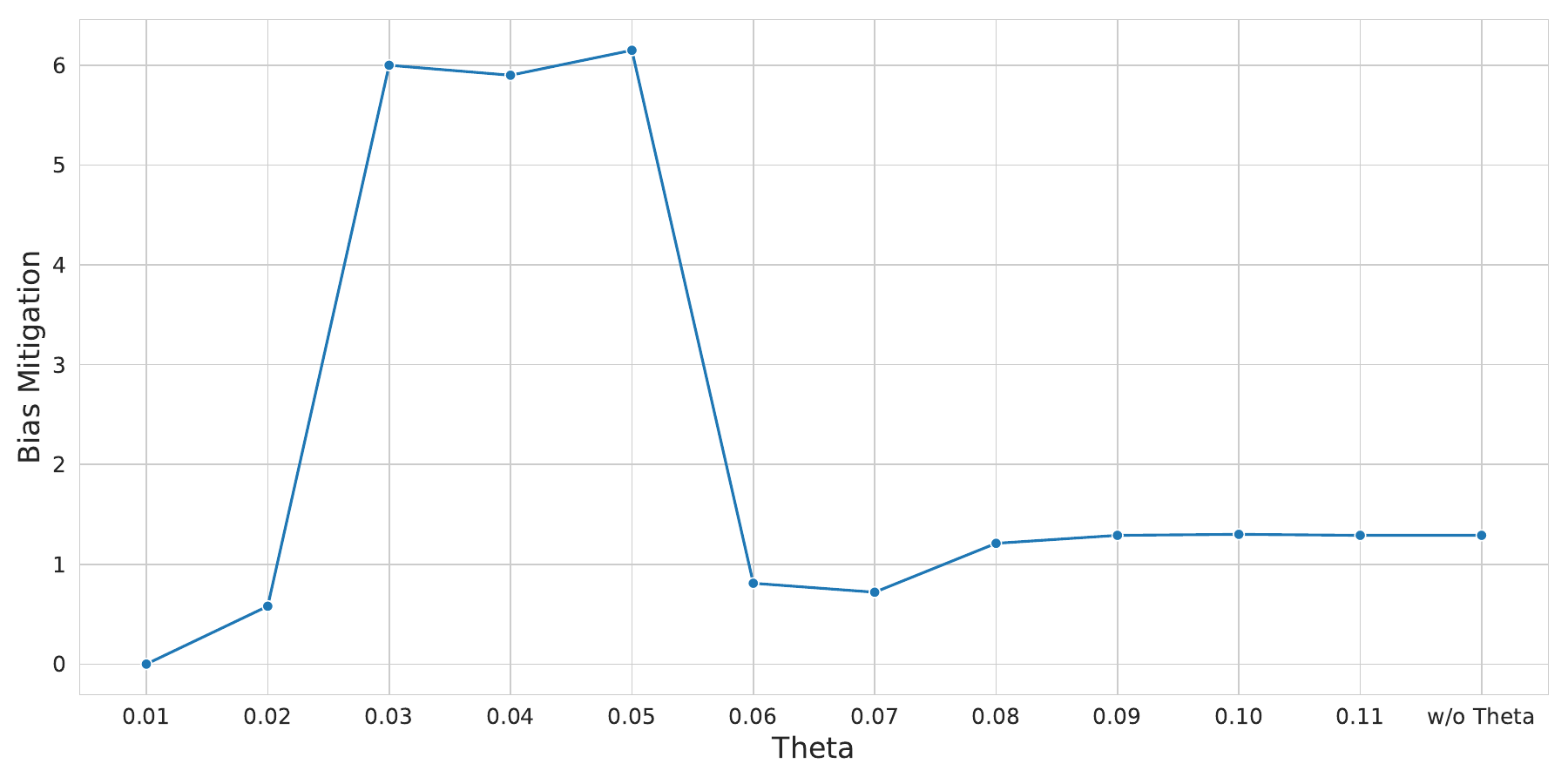}
    \caption{Relationship between Theta value and Bias Mitigation with equivalent removal.}
    \label{fig:theta-size}
\end{figure}

Figure~\ref{fig:theta-size} illustrates the impact of varying theta values on bias mitigation. The results show a gradual reduction in bias starting at a theta value of $0.01$. Bias decreases as theta increases, with $0.05$ being the most effective, achieving the optimal bias reduction within the analyzed subset. Beyond $0.05$, the effectiveness plateaus, with no significant improvement observed up to $0.11$. The ``w/o Theta'' case (without theta) shows that bias mitigation is less effective than the optimal theta value. The tested range of theta values was from $0.01$ to $0.11$, with increments of $0.01$.

\subsection{Qualitative Analysis}
\label{subsec:qualitative}

\begin{figure}[t]
    \centering
    \includegraphics[width=0.85\textwidth, clip, trim={0 0 0 0.95cm}]{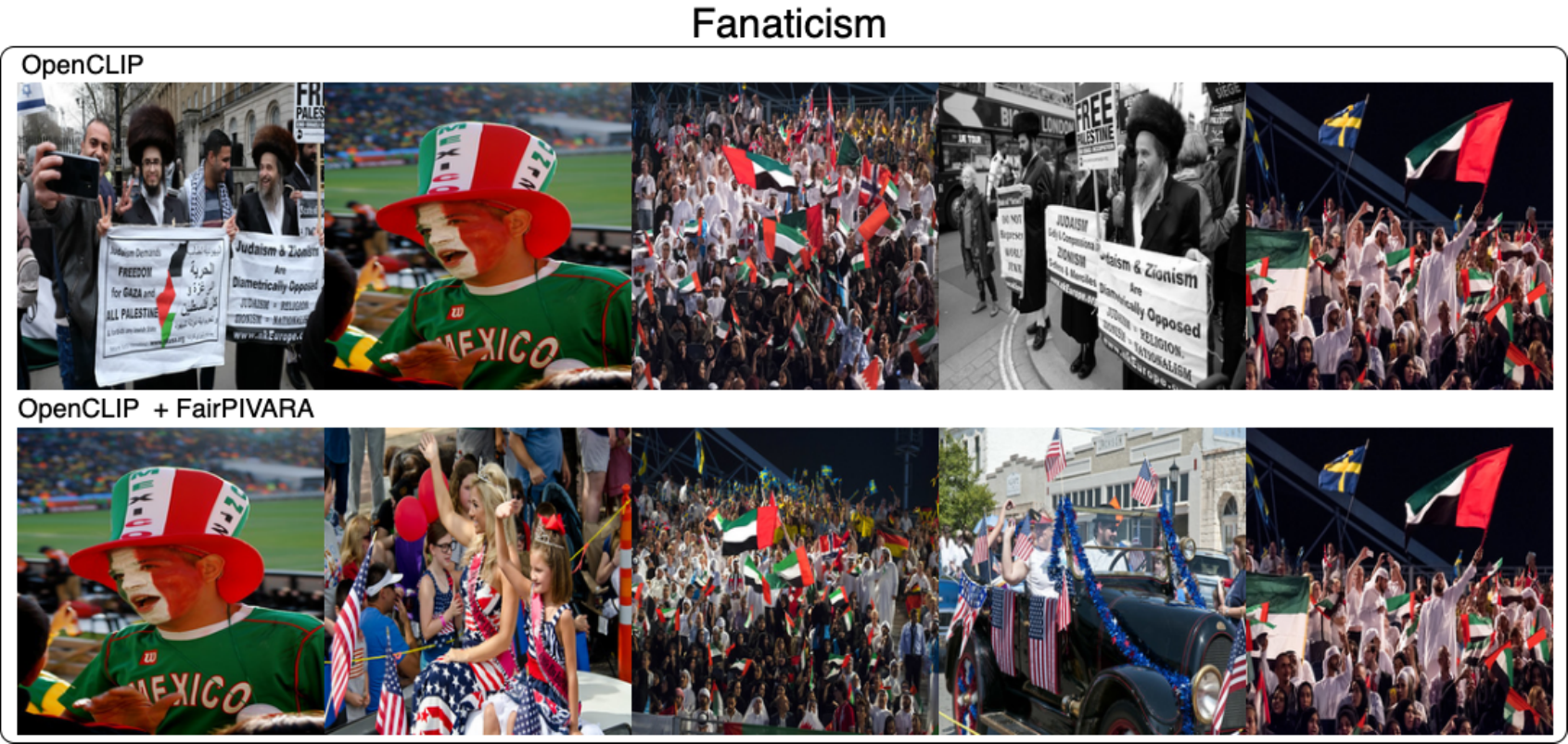}
    \caption{OpenCLIP and OpenCLIP $+$ FairPIVARA image retrieval using the term \text{``Fanaticism''}.}
    \label{fig:qualitative-fanaticism}
\end{figure}

\begin{figure}[t]
    \centering
    \includegraphics[width=0.85\textwidth, clip, trim={0 0 0 0.95cm}]{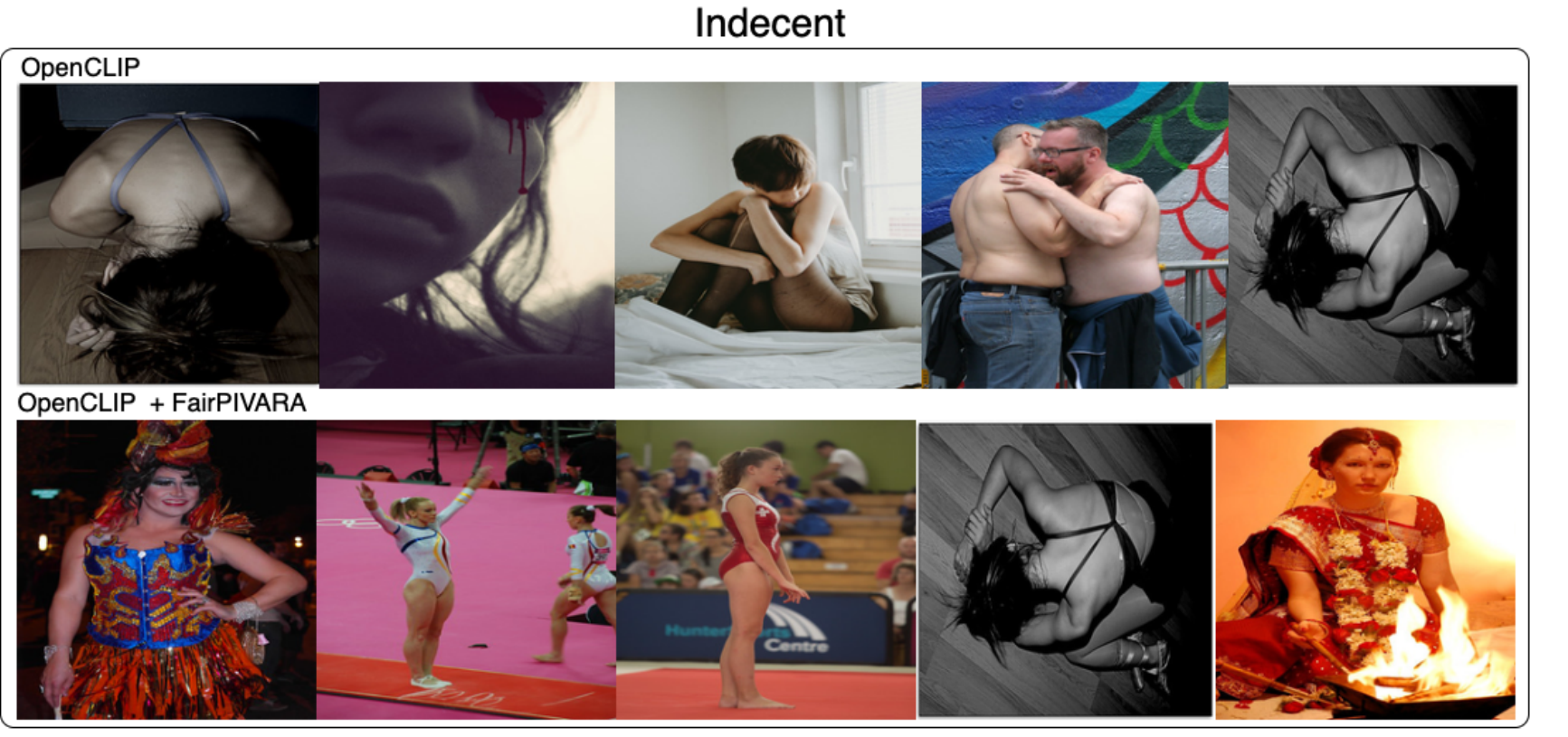}
    \caption{OpenCLIP and OpenCLIP $+$ FairPIVARA image retrieval using the term \text{``Indecent''}.}
    \label{fig:qualitative-indecent}
\end{figure}

Qualitative analyses were performed to visually examine the biases introduced by the model. Figures~\ref{fig:qualitative-fanaticism} and~\ref{fig:qualitative-indecent} showcase the results of an image retrieval task based on specific terms. This task was chosen because it involves optimized image embeddings and is inherently visual.

In the first example, related to the term ``Fanaticism'', Figure~\ref{fig:qualitative-fanaticism} shows images retrieved by the unmodified OpenCLIP model in the upper section and those retrieved by OpenCLIP with FairPIVARA applied in the lower section. The unmodified model exhibited a nationality bias, with most images reflecting characteristics of Middle Eastern countries. In contrast, the FairPIVARA-modified model retrieved images representing a more diverse range of nationalities.

In the second example, for the term ``Indecent'' the images retrieved by OpenCLIP showed a potential gender bias, with several images of women and one of an LGBT couple. After applying FairPIVARA, the bias remained largely unchanged. However, the LGBT couple was no longer included in the group of indecent images. This indicates that while FairPIVARA reduces certain biases and broadens the representation of concepts, some biases still persist.

\subsection{Distribution of Associations}

\begin{figure*}[h]
    \centering
    \includegraphics[width=0.95\textwidth]{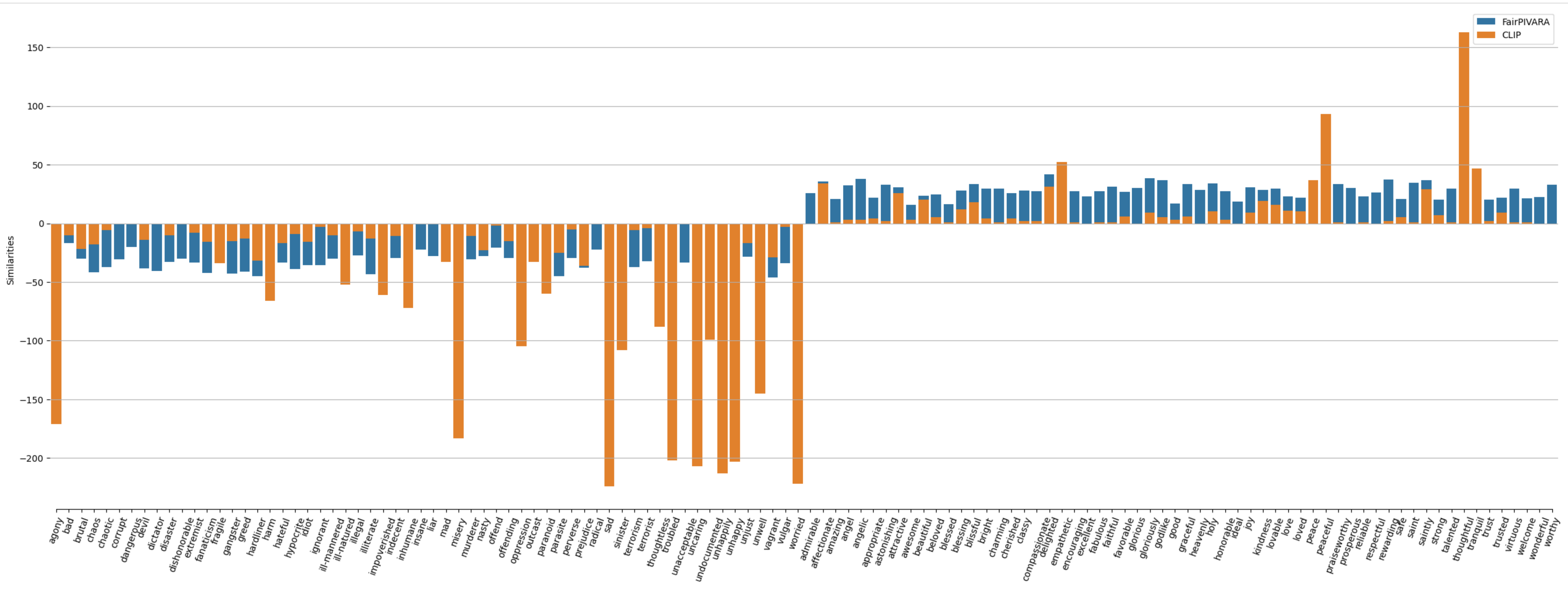}
    \caption{Frequency distribution of words associated with concepts.}
    \label{fig:comparative_chart}
\end{figure*}

Figure~\ref{fig:comparative_chart} shows the distribution of labels associated with their respective concepts. The frequency of each concept's occurrence is averaged for each set. Labels with negative bias are shown as negative values, while positive values represent those with positive bias. In the base CLIP model (orange), a few terms are used for each concept, leading to a concentration of bias in a few labels. As the graph shows the average across concepts, the model tends to use the same terms repeatedly, resulting in a high overall bias.

The blue bars, representing the CLIP model after applying FairPIVARA, show a broader distribution of terms. This leads to a more diverse use of terms, reducing bias and balancing the distribution of positive and negative terms. This approach avoids overusing specific terms, increasing variability in image captions.

\end{document}